\documentclass[final,3p,times]{elsarticle}

\pdfoutput=1
\usepackage{amssymb}
\usepackage{color}
\usepackage{multirow}
\usepackage{amsthm}
\usepackage{amsmath}
\usepackage{booktabs}
\usepackage{tabto}
\usepackage{hyperref}

\usepackage{rotating}
\usepackage{bbding}
\usepackage{wasysym}
\usepackage[linesnumbered,algoruled,boxed,lined]{algorithm2e}

\journal{arXiv}

\begin{document}

\begin{frontmatter}



\title{A pipeline for fair comparison of graph neural networks in node classification tasks}


\author[label1,label2]{Wentao Zhao \corref{cor1}}
\ead{zhaowentao163@163.com}
\author[label3]{Dalin Zhou}
\author[label1]{Xinguo Qiu}
\author[label1]{Wei Jiang}

\affiliation[label1]{organization={College of Mechanical Engineering},
            addressline={Zhejiang University of Technology}, 
            city={Hangzhou},
            postcode={310023}, 
            state={Zhejiang},
            country={China}}
\affiliation[label2]{organization={School of Intelligent Transportation},
      	addressline={Zhejiang Institute of Mechanical \& Electrical Engineering}, 
      	city={Hangzhou},
      	postcode={310053}, 
      	state={Zhejiang},
      	country={China}}
\affiliation[label3]{organization={School of Computing},
	addressline={University of Portsmouth}, 
	city={Portsmouth},
	postcode={PO1 3HE}, 
	state={Hampshire},
	country={UK}}
\cortext[cor1]{Corresponding author}
\fntext[cor2]{\small{\url{https://github.com/karl-zhao/benchmarking-gnns-pyg}}}
\begin{abstract}
Graph neural networks (GNNs) have been investigated for potential applicability in multiple fields that employ graph data. 
However, there are no standard training settings to ensure fair comparisons among new methods, including different model architectures and data augmentation techniques.
We introduce a standard, reproducible benchmark to which the same training settings can be applied for node classification. For this benchmark, we constructed 9 datasets, including both small- and medium-scale datasets from different fields, and 7 different models. We design a k-fold model assessment strategy for small datasets and a standard set of model training procedures for all datasets, enabling a standard experimental pipeline for GNNs to help ensure fair model architecture comparisons. We use node2vec and Laplacian eigenvectors to perform data augmentation to investigate how input features affect the performance of the models.
We find topological information is important for node classification tasks.
Increasing the number of model layers does not improve the performance except on the PATTERN and CLUSTER datasets, in which the graphs are not connected.
Data augmentation is highly useful, especially using node2vec in the baseline, resulting in a substantial baseline performance improvement.
\end{abstract}
%


\end{frontmatter}


\section{Introduction}
Many classical models have been applied to perform deep learning studies on data analysis. The convolutional neural network (CNN) and recurrent neural network (RNN) have been used extensively in computer vision (CV) and natural language processing (NLP), respectively. However, when faced with graph-structured data, such as social networking service (SNS) datasets, knowledge graphs, molecular structures, and biological and financial networks, CNNs and RNNs are not easily applied due to the graph structure of the data~\cite{hamiltonRepresentationLearningGraphs2017}. To solve problems such as node classification~\cite{bhagatNodeClassificationSocial2011}~\cite{zhangEndtoendDeepLearning2018}, graph classification~\cite{yingHierarchicalGraphRepresentation2018a}~\cite{cangeaSparseHierarchicalGraph2018}~\cite{bianchiGraphNeuralNetworks2019}, social recommendation~\cite{freemanVisualizingSocialNetworks2000}~\cite{perozziDeepwalkOnlineLearning2014} and link prediction~\cite{liben-nowellLinkpredictionProblemSocial2007}~\cite{yangEmbeddingEntitiesRelations2014}, graph convolutional network (GCN) models~\cite{kipfSemisupervisedClassificationGraph2016}, which exploit message-passing (or equivalently, various neighborhood functions) to precise high levels, have been proposed.
More works have been done both optimizes performance of GCNs and
solve real life problems.~\cite{zhengGSSAPayAttention2020} optimizes GCNs from the feature importance perspective via statistical self-attention.~\cite{zhaoInfGCNIdentifyingInfluential2020a} put forward a deep learning model, named InfGCN, to identify the most influential nodes in a complex network based on Graph Convolutional Networks.~\cite{guoShorttermTrafficSpeed2020} use graph attention temporal convolutional networks to forecast short-term traffic speed.~\cite{wangMultimodalGraphConvolutional2020} propose a Multimodal Graph Convolutional Networks (MGCN) automatically filter high quality content from a large number of multimedia articles.

Thanks to the establishment of ImageNet~\cite{dengImagenetLargescaleHierarchical2009a}, image classification in computer vision has undergone significant development, and many challenging and realistic benchmark datasets have been proposed that facilitate scalable, robust, and reproducible graph machine learning research.
The Open Graph Benchmark~\cite{huOpenGraphBenchmark2020a} is one recent benchmark initiative whose goal is to represent large real-world datasets from various domains; 
it particularly emphasizes out-of-distribution rationalization performance
through valid data splits. A steady trend toward improvement in model accuracy also has been demonstrated on various benchmarks.
The accuracy of deep CNNs has developed steadily based on both model architecture improvements and data augmentation~\cite{heBagTricksImage2019}. For graph neural networks (GNNs), ~\cite{shchurPitfallsGraphNeural2018a} showed that different data splits lead to different model rankings on node classification tasks, and ~\cite{erricaFairComparisonGraph2019a} proposed a fair performance comparison method for GNN architectures. Inspired by the above studies, we constructed a standardized and reproducible experimental environment in this study and used it to test how model architectures and data augmentation techniques impact node classification accuracy.
The main contributions of our work are as follows.

\begin{itemize}
\item	We collect a total of 9 datasets~\cite{wuComprehensiveSurveyGraph2020}~\cite{dwivediBenchmarkingGraphNeural2020a} and 7 models that are widely used in various fields in the PyTorch~\cite{paszkePytorchImperativeStyle2019} and PyTorch geometric~\cite{feyFastGraphRepresentation2019} frameworks and use them to compare the expressive power of various model structures on different node classification tasks. For fair comparisons, we use 2 parameter budgets to ensure that the number of parameters and layers is roughly the same. To conduct further research, researchers can easily extend our framework by adding new models with different features and arbitrary datasets from their own experiments.
\item	We propose a method for fair comparison using the small datasets to evaluate model performance. The existing prior results show that applying small datasets with specific training/evaluation/test splits is unsuitable for making fair model comparisons. Therefore, we specify a 10-fold cross validation split for the small datasets instead of randomly splitting the data several times. We use this approach to build model assessment methods. We perform each experiment several times under the same training conditions to ensure fair comparisons.
\item	We test some input features to assess the expressive power of embedding by using node2vec and Laplacian vectors as data augmentation techniques to represent the node position and add to the input features. Using this approach, we can evaluate how the various input features affect the results.
\end{itemize}

The goal of this study is not to find the best-performing GNNs or to obtain the highest classification accuracy (which is computationally expensive) in the specific targets using an extensive hyperparameter grid search. Instead, our goal is to construct a standard benchmark for node classification and a uniform evaluation framework for GNNs. We use numerous GNNs to evaluate the power of the proposed architecture, and future researchers can easily use the framework to compare new proposed models using the existing architectures.

\section{Training Procedures}
A training process includes mainly datasets, models, and the strategy for updating model parameters. As we can see in the baseline training procedure in Figure~\ref{fig1}, Algorithm~\ref{alg1} performs node classification with many graphs, while Algorithm~\ref{alg2} performs node classification for only one graph. For node classification, $X$ is the embedding of the batch graph whose dimension is $N\times d$, where $N$ denotes the number of nodes in the batch graphs,  $d$ denotes the dimension of each node, $y$ is the node label, and the overall dimension is $N\times d$
, where $N$ denotes the number of nodes in the batch graphs and $d$ denotes the dimension of the node label.

The difference between the 2 algorithms is that node classification from only one graph does not require a batch because for some node classification tasks, only one graph is available, and it is difficult to separate 
the nodes 
due to their relations. We emphasize that not all node classifications involve only one graph; some have multiple graphs and use minibatches during training. For these tasks, the algorithm process used is similar to Algorithm~\ref{alg1}. For datasets that include more nodes and edges, it is easy to exceed the CUDA memory limitations when training models on a GPU; consequently, various methods have been proposed to solve the problems. One is to use a clustering algorithm to separate the graph into clusters and then randomly select some clusters for training, similar to algorithm ~\ref{alg1}~\cite{chiangClusterGCNEfficientAlgorithm2019}. The other approach is to use sample subgraphs obtained via a random walk sampling algorithm~\cite{zengGraphsaintGraphSampling2019}.

\begin{figure}[!ht]
	\centering
	\begin{minipage}{0.8\textwidth}
		\IncMargin{1em}
		\begin{algorithm}[H]
			\caption{Training a neural network with the Adam optimizer for node classification}
			\label{alg1}
			\SetKwData{Left}{left}\SetKwData{This}{this}\SetKwData{Up}{up}
			\SetKwFunction{Union}{Union}\SetKwFunction{FindCompress}{FindCompress}
			\SetKwInOut{Input}{Input}\SetKwInOut{Output}{Output}
			
			\Input{Node embedding for every graph, initialize(net)}
			\Output{Trained model}
			\BlankLine
			\For{epoch = 1,2,...,K}{
				\For{batch = 1,2,...,\#graphs / b }{
					graphs$\leftarrow$  uniformly random sample \b graphs\;
					X,y $\leftarrow$  preprocess(graphs)\;
					z $\leftarrow$  forward(net,X)\;
					l $\leftarrow$  loss(z, y)\;
					grad $\leftarrow$  backward(l)\;
					update(net,grad)
			}}
		\end{algorithm}
		\DecMargin{1em}
	\end{minipage}
	
	\begin{minipage}{0.8\textwidth}
		\IncMargin{1em}
		\begin{algorithm}[H]
			\caption{Training a neural network with the Adam optimizer for node classification}
			\label{alg2}
			\SetKwData{Left}{left}\SetKwData{This}{this}\SetKwData{Up}{up}
			\SetKwFunction{Union}{Union}\SetKwFunction{FindCompress}{FindCompress}
			\SetKwInOut{Input}{Input}\SetKwInOut{Output}{Output}
			\Input{Node embedding for every graph, initialize(net)}
			\Output{Trained model}
			\BlankLine
			\For{epoch = 1,2,...,K}{
				X,y $\leftarrow$  preprocess(graphs)\;
				z $\leftarrow$  forward(net,X)\;
				l $\leftarrow$  loss(z, y)\;
				grad $\leftarrow$  backward(l)\;
				update(net,grad)
			}
		\end{algorithm}
		\DecMargin{1em}
	\end{minipage}
	\caption{A baseline training procedure for node classification}
	\label{fig1}
\end{figure}

\subsection{Datasets}\label{sec2.1}
The datasets used in our experiments mainly stem from studies~\cite{dwivediBenchmarkingGraphNeural2020a}~\cite{wuComprehensiveSurveyGraph2020}~\cite{senCollectiveClassificationNetwork2008} that make benchmarks for different tasks. Here, we introduce the 9 datasets summarized in Table~\ref{tab1}, including their split schemes, splits, features and metrics. More details (such as the classes, domains, numbers of nodes and edges, and dimensions of the input features) are shown in Table~\ref{tab6} (Appendix A).


\begin{table}[!ht]
	\caption{Summary of the datasets used in the experiments. A split strategy, the prediction task and the input features are described for each dataset. Additional details can be found in Section~\ref{sec2.1}.}
	\label{tab1}
	\centering
	\footnotesize
	\begin{tabular}{cccccccc}
		\toprule
		\textbf{Name}	& \textbf{\#Tasks}	& \textbf{Split   Scheme} & \textbf{Split}	& \textbf{Node Feat.}	& \textbf{Edge Feat.} & \textbf{directed}	& \textbf{Metric}\\
		\midrule
		cora     & 1       & 10-fold CV     & 08:01:01             & \CheckmarkBold         & \XSolidBrush      & \XSolidBrush & Acc     \\
		CiteSeer & 1       & 10-fold CV     & 08:01:01             & \CheckmarkBold               & \XSolidBrush      & \XSolidBrush   & Acc     \\
		PubMed   & 1       & 10-fold CV     & 08:01:01             & \CheckmarkBold       & \XSolidBrush    & \XSolidBrush  & Acc     \\
		PATTERN  & 1       & random         & 05:01:01             & \CheckmarkBold       & \XSolidBrush     & \XSolidBrush  & Acc     \\
		CLUSTER  & 1       & random         & 10:01:01             & \CheckmarkBold      & \XSolidBrush    & \XSolidBrush   & Acc     \\
		Products & 1       & sales rank     & 196615:39323:2213091 & \CheckmarkBold      & \XSolidBrush    & \XSolidBrush  & Acc     \\
		arXiv    & 1       & time           & 90941:29799:48603    & \CheckmarkBold      & \XSolidBrush     & \CheckmarkBold  & Acc     \\
		MAG      & 1       & time           & 629571:64879:41939   & \CheckmarkBold      & \XSolidBrush      & \CheckmarkBold   & Acc     \\
		Proteins & 112     & species        & 86619:21236:24679    & \XSolidBrush       & \CheckmarkBold      & \XSolidBrush   & ROC-AUC\\
		\bottomrule
	\end{tabular}
\end{table}

\textbf{Cora}, \textbf{CiteSeer} and \textbf{PubMed}:
For the node classification tasks, we first executed our experiment on three widely used benchmark datasets: Cora, CiteSeer and PubMed~\cite{senCollectiveClassificationNetwork2008}~\cite{yangRevisitingSemisupervisedLearning2016}, which are all citation networks.

Prediction task:
For the Cora dataset, the task is to predict the subject of the paper (node) based on the surrounding node data and the graph structure. The different subjects are `Case Based', `Genetic Algorithms', `Neural Networks', `Probabilistic Methods', `Reinforcement Learning', `Rule Learning', and `Theory'. For the CiteSeer dataset, the task is to predict which domain the paper (node) belongs to. All the papers are grouped into 6 classes: `Agents', `AI', `DB', `IR', `ML' and `HCI'. The publications from the PubMed dataset pertaining to diabetes are classified into one of three classes: `Diabetes Mellitus, Experimental', `Diabetes Mellitus Type 1', or `Diabetes Mellitus Type 2'.

Dataset splitting:
An earlier work~\cite{shchurPitfallsGraphNeural2018a} compared the performance of various GNNs on node classification tasks and found that choosing different training/validation/testing splits leads to different performance rankings. This result occurs primarily because the datasets used were too small to fully leverage the power of data-hungry GNNs.
For fair comparisons, a 10-fold cross-validation split is used to train the models, similar to reference~\cite{erricaFairComparisonGraph2019a}. Ten sets of training, validation and testing data indices at a ratio of 8:1:1, respectively, are used to perform node classifications and calculate the test accuracy. Figure~\ref{fig2} reports the pseudocode of the entire process used to obtain the results of the classification accuracy, and the overall procedure is visually summarized in Figure~\ref{fig4}. Note that the model training processes are the same for Cora, CiteSeer and PubMed and that the datasets are represented later. We introduce the experimental setting in detail below.

\textbf{PATTERN} and \textbf{CLUSTER} are both artificially generated datasets generated with stochastic block models~\cite{abbeCommunityDetectionStochastic2017}. These datasets are widely used to model communities by modulating intra and extra community connections to control the difficulty of tasks in social networks, and they can be used for node classification tasks. PATTERN and CLUSTER contain a total of 14 K and 12 K graphs, respectively. The distribution of the number of nodes in these graphs is depicted in Figure~\ref{fig5}.

Prediction task:
PATTERN~\cite{scarselliGraphNeuralNetwork2008} is used to recognize specific predetermined subgraphs. For all the data, a graph $G$ is generated with 5 communities whose sizes are randomly selected between [5,35]. We then randomly generate 100 patterns $P$ composed of 20 nodes. The node features for $P$ and $G$ are generated as a random signal with values of {0,1,2}. The graph sizes are 44--195. The output node labels have a value of 1 if the node belongs to $P$ and 0 if it belongs to $G$.
CLUSTER is targeted at identifying community clusters. We generate 6 SBM clusters with randomly selected sizes in the range [5,35]. The graphs contain 40--190 nodes. Each node is embedded with a value from {0,1,2,..,6}. If the value of the node is $a(a>0)$, the node belongs to class $a-1$, and if the value is 0, the node class is unknown. Only one labeled node is randomly assigned to each community, and most of the features are set to 0. When calculating accuracy, we use weighted accuracy w.r.t. the class sizes.

Dataset splitting:
The PATTERN and CLUSTER datasets are both split randomly. The PATTERN dataset split includes 10,000 training, 2,000 validation, and 2,000 testing graphs, while the CLUSTER dataset is split into 10,000 training, 1,000 validation, and 1,000 testing graphs. We save the splits and subsequently use them to ensure fair comparisons.
In addition to the artificially generated datasets, we import some datasets from the open graph benchmark dataset (OGB)~\cite{huOpenGraphBenchmark2020a}. The graphs all stem from real-world tasks. Unlike the datasets discussed above, these datasets include only one graph.

\textbf{Products}:
The Products dataset is an undirected and unweighted graph. It is an Amazon product co-purchasing network~\cite{Bhatia16}. The node features and the edges between 2 nodes are represented by a dimensionality-reduced bag-of-words of the product descriptions and the products that are purchased together, respectively, following ~\cite{chiangClusterGCNEfficientAlgorithm2019}.

Prediction task:
The task is to predict which category the product belongs to. Forty-seven top-level categories are used as the target labels.

Dataset splitting:
The previously used Cora, CiteSeer, and PubMed citation networks are typically split randomly; however, instead of performing a random split, ~\cite{huOpenGraphBenchmark2020a} split the dataset by sales ranking. The top 8\%, the next top 2\% and the remainder of the records are used for training, validation and testing, respectively. Thus, as in real-world scenarios, we label the most popular products and leave the less popular products for validation and testing. This is a more realistic segmentation approach.

\begin{figure}[!ht]
	\centering
	\begin{minipage}{0.8\textwidth}
		\IncMargin{1em}
		\begin{algorithm}[H]
			\caption{Model Assessment($k$-fold CV)}
			\label{alg3}
			\SetKwData{Left}{left}\SetKwData{This}{this}\SetKwData{Up}{up}
			\SetKwFunction{Union}{Union}\SetKwFunction{FindCompress}{FindCompress}
			\SetKwInOut{Input}{Input}\SetKwInOut{Output}{Output}
			
			\Input{Node embedding for graph $\mathcal{D}$, initialize(net)}
			\Output{Test Acc.}
			\BlankLine
			Split $\mathcal{D}$ into k folds $F_1$,...,$F_k$\;
			\For{i = 1,2,...,k}{
				$train_k,valid_k,test_k \leftarrow \left(\cup_{j \neq i} F_{j}\right), F_{i}$\;
				\For{z = 1,2,...,R}{
					$model_r \leftarrow Train(train_k,valid_k)$\;
					$p_r \leftarrow Eval(model_k, test_k)$\;
				}
				$ perf_k \gets \sum_{r=1}^{R} \mathrm{p}_{r} / R$\;
			}
			return $\sum_{i=1}^{k} perf_i / k$
			
		\end{algorithm}
		\DecMargin{1em}
	\end{minipage}
	
	\begin{minipage}{0.8\textwidth}
		\IncMargin{1em}
		\begin{algorithm}[H]
			\caption{Model Training}
			\label{alg4}
			\SetKwData{Left}{left}\SetKwData{This}{this}\SetKwData{Up}{up}
			\SetKwFunction{Union}{Union}\SetKwFunction{FindCompress}{FindCompress}
			\SetKwInOut{Input}{Input}\SetKwInOut{Output}{Output}
			
			\Input{$train_k,valid_k, initialize(net)$}
			\Output{Trained model}
			\BlankLine
			\For{epoch = 1,2,...,N}{
				$model \leftarrow Train(train_k)$\;
				$lossval_n \leftarrow Eval(model,valid_k)$\;
				\While{$lossval_n > lossval_{n-1}$}{
					$patience++ $\;
					\If{$ patience>=lr\_schedule\_patience$ }{
						$lr = lr \times 0.5$\;
						$patience = 0$\;
				}}
				\If{$lr <= min_{lr}$ \textbf{or} $time>max_{time}$}{
					$ break$}
			}
			return model
		\end{algorithm}
		\DecMargin{1em}
	\end{minipage}
	\caption{Pseudocode for model assessment (top) and model training (bottom). In Algorithm 3, `Train' refers to Algorithm 4, whereas a strategy is introduced to train the models according to the results of the inference phases. The models are then used to test the classification accuracy in the $test_k$ part and obtain the $p_r$. We train the models and test them $R$ times and obtain the average $perf_k$. Finally, the experiments are used in the k-fold CV, and the average results are obtained.}
	\label{fig2}
\end{figure}

\textbf{arXiv}:
arXiv is a directed graph that represents the citation network between all computer science (CS) arXiv papers~\cite{wangMicrosoftAcademicGraph2020}. The nodes represent an arXiv paper embedded with a 128-dimensional feature vector obtained by averaging the embeddings of the words in its title and abstract.

Prediction task:
The task is to predict the subject areas to which an arXiv CS paper belongs (e.g., cs.AI, cs.LG, and cs.OS). The dataset includes forty labels that were manually determined by the paper's authors and arXiv moderators.

Dataset splitting:
We use a realistic data split based on the publication date of the papers to split this dataset; thus, papers published through 2017 are used for training, and papers published in 2018 and 2019 are used as validation and testing datasets, respectively.

\textbf{MAG}:
MAG is a heterogeneous network composed of a subset of the Microsoft Academic Graph (MAG)~\cite{wangMicrosoftAcademicGraph2020}. It consists of 4 entity types: papers (736,839 nodes), authors (1,134,649 nodes), fields of study (59,965 nodes) and institutions (8,740 nodes). In addition, 4 types of directed relations connect 2 of the entity types. Each author is affiliated with an institution, an author writes a paper, the paper cites other papers, and each paper has a primary topic field of study. A 128-dimensional word2vec feature vector is used as the paper node embeddings. The other node types are not associated with the input node features.

Prediction task:
The task is to predict the venue (conference or journal) at which each paper was presented. Based on the data, the papers could theoretically have been presented at any of 349 different venues, making the task a 349-class classification problem. To simplify the task, we consider only the paper nodes and the types of directed relations connecting 2 entity types, namely, a paper citing another paper, to transfer the task to be dealt with by an isomorphic graph.

Dataset splitting:
Similar to arXiv data, we consider a realistic data split based on the paper publication dates. In practice, we always train on older existing papers and test on newer published papers. Finally, we used the papers published through 2017 for training. The papers published in 2018 and 2019 are used as the validation and testing datasets, respectively.

\textbf{Proteins}:
Proteins consist of undirected, weighted, and typed graphs where the nodes represent proteins, and the edges indicate different types of biologically meaningful associations between proteins~\cite{szklarczykSTRINGV11Protein2019}~\cite{consortiumGeneOntologyResource2019}. All the edges are embedded with 8-dimensional features. Each dimension represents the strength of a single association type and takes a value between 0 and 1. The dataset includes a total of 8 association types, constituting 8-dimensional features.
In contrast to the other datasets, the proteins do not have node features; they possess only edge features---but there are more than 30 million edges. To better train the model using the input features, we use the averaged edge features of incoming edges as node features.

Prediction task:
The task is to predict whether various protein functions are present. In total, there are 112 possible labels to predict. The result is measured by the average of the ROC-AUC scores across the 112 tasks.

Dataset splitting:
We split the protein nodes according to the species from which the proteins come. This makes it possible to evaluate model generalizability across different species.

\subsection{Embedded Information Regarding the Datasets}
Using Laplacian eigenvectors as positional embeddings: For a graph, the positional features that represent each node are important, especially when working with graphs that exhibit symmetry in their structures, such as node or edge isomorphism~\cite{murphyRelationalPoolingGraph2019}~\cite{srinivasanEquivalenceNodeEmbeddings2019}. Laplacian eigenvectors~\cite{belkinLaplacianEigenmapsDimensionality2003} are a technique that embeds graphs into Euclidean space. The eigenvectors are used to build a meaningful local coordinate system that preserves the global graph structure. The eigenvectors are defined via factorization of the graph's Laplacian matrix.

\begin{equation}
\Delta=\mathrm{I}-D^{-1 / 2} A D^{-1 / 2}=U^{T} \Lambda U
\end{equation}
The Laplacian eigenvectors of a graph can be calculated by Laplace decomposition, where $A$ is the $n \times n$ adjacency matrix, $D$ is the degree matrix, $\Lambda$ is the eigenvalue and $U$ is the eigenvector. $U$ is a type of positional encoding; the $k$-smallest nontrivial encoding is used as a supplement to the node embeddings to enrich the input node information. Laplacian eigenvectors are used to provide smooth encoding coordinates for neighboring nodes.

Node2vec~\cite{groverNode2vecScalableFeature2016}:
In NLP tasks, word2vec~\cite{mikolovEfficientEstimationWord2013} is a commonly used word embedding method that describes the cooccurrence relationships between words in sentence sequences in the corpus and then learns the vector representation of words. The idea behind node2vec is similar to word2vec; it uses the node co-occurrences in the graph to learn the vector representations of nodes. In contrast to deepwalk~\cite{perozziDeepwalkOnlineLearning2014} or randomwalk to obtain the nearest neighbor sequence of vertices, node2vec uses a biased random walk. It is an unsupervised node embedding method.

\subsection{Model Selection}\label{sec2.3}
\begin{figure}[!ht]
	\centering
	\includegraphics[width=15.6 cm]{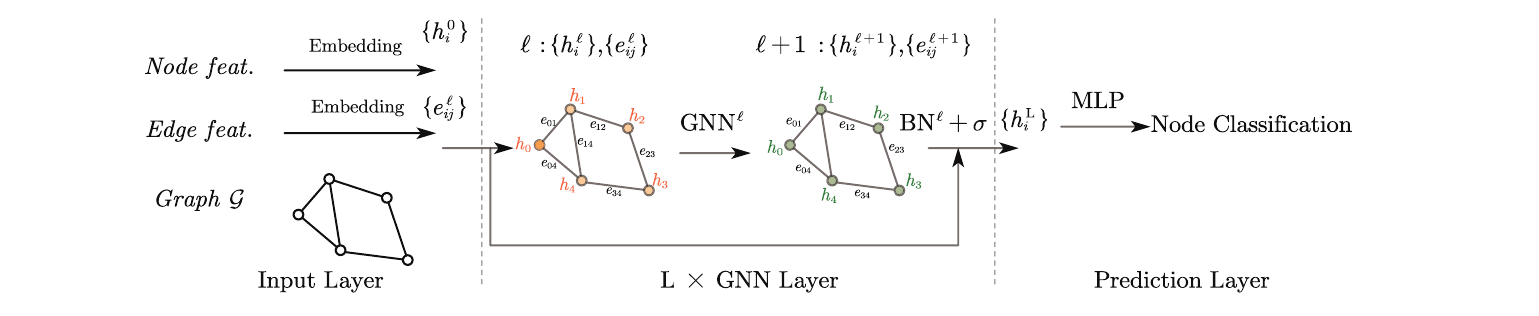}
	\caption{A standard experimental pipeline for GNNs that embeds the graph nodes and edges to dimensions of $N \times hidden\_dim $ and $E \times hidden\_dim $, respectively, and then uses them as input features with the $edge\_index$ in the input layer. L layers of the GNN layers are used with the batch norms, activation functions and residual connections. Finally, an MLP is used to embed the nodes with $N \times hidden\_dim $ to $N \times classes $ to perform classification.}
	\label{fig3}
\end{figure}

To ensure fair model comparisons, a parameter budget for the models is controlled in our experiments. The pipeline of experiments is illustrated in Figure~\ref{fig3} for all GNNs. First, an embedding layer is used in the input layer to satisfy the input requirements of the GNN layers. GNN layers are then used in the $L \times GNN$ layers as well as the batch norm and activation function. Finally, MLP is used to embed the graph.

Most GNN implementations involve a GCN-Graph Convolutional Network~\cite{kipfSemisupervisedClassificationGraph2016}, GAT-Graph Attention Network~\cite{velickovicGraphAttentionNetworks2017}, GraphSAGE~\cite{hamiltonInductiveRepresentationLearning2017}, GIN--Graph Isomorphism Network~\cite{xuHowPowerfulAre2018}, MoNet--Gaussian Mixture Model Network~\cite{montiGeometricDeepLearning2017}, Residual Gated Graph Convolutional Network~\cite{bressonResidualGatedGraph2017}, and Gated Graph Convolutional Network~\cite{liGatedGraphSequence2015}. All these model architectures are explained in~\ref{appB}.

\section{Experiments}\label{sec3}
\unskip
\subsection{Data Splitting and Model Assessment}
On small datasets such as Cora, CiteSeer and PubMed, we use the $k$-fold cross-validation split for model assessment. As shown in Algorithm~\ref{alg3} (Figure~\ref{fig2}), we first split the datasets into $k$ folds ($k$ = 10). Note that the class proportions are preserved in all the data splits (training, validation, and testing splits). All the data partitions are preprocessed; thus, the models are trained and evaluated on the same data splits.
We train the models $R$ times ($R$= 3) on each training and validation fold and obtain an average accuracy $perf_k$ for each fold. The average of the $k$ folds is used as the final accuracy result of the classification. Finally, we execute each model 4 times with different seeds and use the mean value as the final classification result. On the other datasets, the respective strategies for splitting the datasets are introduced above; each model is executed 4 times, and the mean accuracy is taken as the final classification result.
\subsection{Training}
The goal of this study is not to find an optimal set of hyperparameters to obtain the best model performance but rather to compare the power of models within a parameter budget.
For the same datasets, we use the same hyperparameters to train the models for node classification. A budget of parameters---including different numbers of layers---is used to help ensure a fair model comparison; the details can be seen in the results.
The model training process strategy is the same for all the datasets, and the value of $N$ is set to 1,000 (Algorithm~\ref{alg4} in Figure~\ref{fig2}). If the inference results $lossval_n$ are not decreased lr\_schedule\_patience times, the learning rate will decrease by half. We implement early stopping when $lr<=min_{lr}$ or $time > max_{time}$. For the loss function, all use the CrossEntropyLoss except the proteins, which use the BCEWithLogitsLoss because of the special classified content. The Adam optimizer~\cite{kingmaAdamMethodStochastic2014} for all models is used in the experiments, and the hyperparameters, e.g., initial learning rate, patience times, $min_{lr}$, and $max_{lr}$, are shown in the code.

\section{Results and Discussion}
We perform the experiments on 3 planetoid datasets, Cora, CiteSeer and PubMed. The model training and model assessment methods are shown in Section~\ref{sec3} (Experiments), and each dataset has 2 parameter budgets. Due to the different dimensions of the input features, different numbers of parameter budgets are used for each dataset. For Cora, CiteSeer and PubMed, 260k/460k, 550k/750k and 150k/350k parameters are used, respectively. The small parameter budget has 4 layers, while the larger budget has 16 layers.
The experimental pipeline for GNNs is shown in Section~\ref{sec2.3}(Model selection). Our goal is to conduct a fair comparison of the different model architectures on small datasets using the same settings.

\begin{table}[!ht]
	\caption{Performance on the Planetoid datasets with 10-fold cross-validation. All the experiments were executed 4 times using the same hyperparameters with different seeds, and the mean ± standard deviation of the results was calculated. The highest value and good values are indicated in \textcolor[rgb]{ 1,  0,  0}{red} and \textcolor[rgb]{ 0,  .439,  .753}{blue} respectively.}
	\label{tab2}
	\centering
	\footnotesize
	\begin{tabular}{cccccccc}
		\toprule
		\textbf{dataset} & \textbf{Model}	& \textbf{L}	& \textbf{params} & \textbf{Train Acc.}	& \textbf{Val Acc.}	& \textbf{Test Acc.} & \textbf{Epoch}\\
		\midrule
		\multirow{16}{*}{\rotatebox{90}{Cora}} 	&	\multirow{2}{*}{MLP}    & 4  & 260167 & 90.873±5.5075  & 54.4803±2.4206 & 53.6274±2.8318 & 217.9±3.2939    \\
		&& 16 & 458311 & 46.8908±2.2576 & 39.1298±1.1797 & 38.2718±1.0889 & 222.8833±0.8813 \\
		&\multirow{2}{*}{GCN}	& 4  & 261191 & 99.9996±0.0008 & 85.4674±0.1454 & 84.9894±0.119  & 225.6583±0.9539 \\
		&& 16 & 462407 & 99.9604±0.0092 & 85.5166±0.2035 & 84.6536±0.191  & 226.55±1.9809   \\
		&\multirow{2}{*}{GraphSage}	& 4  & 263673 & 100.0±0.0      & 85.9472±0.2164 & 85.515±0.2481  & 216.2333±0.9416 \\
		&& 16 & 466625 & 100.0±0.0      & 86.3684±0.4295 & 85.6969±0.3015 & 211.2334±0.2108 \\
		&\multirow{2}{*}{ResGatedGCN}	& 4  & 264387 & 99.9996±0.0008 & 85.9225±0.4375 & 85.3889±0.6281 & 222.2583±0.9492 \\
		&& 16 & 467887 & 99.9973±0.0054 & 86.1562±0.2771 & 85.7521±0.3085 & 216.9083±1.113 \\
		&\multirow{2}{*}{GatedGCN}	 & 4  & 268739 & 99.9642±0.017  & 86.3377±0.5164 & \textcolor[rgb]{ 0,  .439,  .753}{86.2045±0.341}  & 237.75±2.0314   \\
		&& 16 & 465219 & 98.1284±0.1248 & 85.8856±0.2054 & 84.8385±0.2387 & 382.7583±2.8856 \\
		&\multirow{2}{*}{GAT}    & 4  & 262215 & 99.9907±0.0045 & 85.5658±0.0714 & 85.1954±0.1021 & 215.825±0.5494  \\
		&& 16 & 466503 & 99.9569±0.0078 & 85.981±0.1752  & 85.1492±0.2009 & 213.5167±2.0943 \\
		&\multirow{2}{*}{GIN}	 & 4  & 261069 & 100.0±0.0      & 87.0388±0.1481 & \textcolor[rgb]{ 0,  .439,  .753}{86.7398±0.0972} & 214.0416±0.1642 \\
		&& 16 & 461176 & 99.9919±0.0051 & 87.0541±0.2471 & \color{red}{87.0323±0.1584} & 225.1333±0.7488 \\
		&\multirow{2}{*}{Monet}   & 4  & 265080 & 100.0±0.0      & 85.6796±0.3681 & 85.4138±0.3714 & 233.1334±2.6961 \\
		&& 16 & 464012 & 100.0±0.0      & 85.6243±0.491  & 85.287±0.5702  & 229.3667±3.1476 \\
		\midrule
		\multirow{16}{*}{\rotatebox{90}{CiteSeer}}	&	\multirow{2}{*}{MLP}   & 4  & 550694 & 98.1854±1.1199 & 56.1987±2.3063 & 55.3595±2.2809 & 212.0±0.9518    \\
		&& 16 & 748838 & 51.4018±1.2994 & 35.4454±0.6745 & 35.347±0.8122  & 222.675±1.5597  \\
		&\multirow{2}{*}{GCN}	 & 4  & 551718 & 99.9471±0.0041 & 70.3678±0.045  & 70.3118±0.0943 & 214.325±0.7416  \\
		&& 16 & 752934 & 99.7125±0.0565 & 70.5005±0.2185 & 70.7902±0.196  & 207.1916±6.2345 \\
		&\multirow{2}{*}{GraphSage}	& 4  & 547381 & 99.9775±0.0    & 72.8328±0.0484 & \textcolor[rgb]{ 0,  .439,  .753}{72.6895±0.1979} & 208.6083±0.25   \\
		&& 16 & 753812 & 99.9775±0.0    & 72.002±0.3888  & 71.9955±0.3202 & 201.7666±0.9031 \\
		&\multirow{2}{*}{ResGatedGCN}   & 4  & 549534 & 99.9775±0.0    & 72.7853±0.7475 & \textcolor[rgb]{ 0,  .439,  .753}{72.662±0.7147}  & 211.5666±0.6492 \\
		&& 16 & 758467 & 99.9775±0.0    & 72.5751±0.4462 & \textcolor[rgb]{ 0,  .439,  .753}{72.5316±0.3692} & 205.7417±0.3304 \\
		&\multirow{2}{*}{GatedGCN}	 & 4  & 549282 & 99.975±0.0051  & 73.1131±0.1474 & \textcolor[rgb]{ 0,  .439,  .753}{72.3867±0.0525} & 221.5084±0.747  \\
		&& 16 & 753285 & 99.5845±0.0472 & 72.44±0.415    & 71.4648±0.2711 & 267.825±0.3696  \\
		&\multirow{2}{*}{GAT}   & 4  & 552742 & 99.9236±0.0076 & 70.4905±0.1094 & 70.9206±0.1353 & 206.3583±0.5209 \\
		&& 16 & 757030 & 99.7893±0.0195 & 70.7558±0.2116 & 71.091±0.1623  & 199.875±0.9472  \\
		&\multirow{2}{*}{GIN}	 & 4  & 549232 & 99.9743±0.0016 & 73.3083±0.1784 & \textcolor[rgb]{ 0,  .439,  .753}{73.9664±0.0609} & 206.0666±0.4422 \\
		&& 16 & 752038 & 99.964±0.0048  & 74.024±0.0876  & \color{red}{74.5226±0.0924} & 210.4±0.2341    \\
		&\multirow{2}{*}{Monet}   & 4  & 552277 & 99.9775±0.0    & 70.7908±0.5502 & 70.6258±0.4377 & 218.1166±0.5815 \\
		&& 16 & 757010 & 99.9775±0.0    & 70.3879±0.8566 & 70.0163±0.9597 & 211.6±3.5057    \\
		\midrule
		\multirow{16}{*}{\rotatebox{90}{PubMed}}	&		\multirow{2}{*}{MLP}    & 4  & 152533 & 93.4437±0.0901  & 87.1062±0.1012  & \textcolor[rgb]{ 0,  .439,  .753}{87.4948±0.1509}  & 364.7±4.791     \\
		&& 16 & 348336 & 75.1723±14.5209 & 71.9386±13.2933 & 72.1064±13.3417 & 372.7±80.8246   \\
		&\multirow{2}{*}{GCN}	& 4  & 145016 & 99.9174±0.0169  & 86.7661±0.0859  & \textcolor[rgb]{ 0,  .439,  .753}{87.1989±0.0994 } & 254.8334±2.3816 \\
		&& 16 & 352496 & 99.7014±0.112   & 86.052±0.0587   & 86.2949±0.0843  & 247.3±5.5586    \\
		&\multirow{2}{*}{GraphSage}	& 4  & 151422 & 99.3514±0.1177  & 86.341±0.1768   & 86.3776±0.1825  & 254.375±0.5725  \\
		&& 16 & 353523 & 99.7196±0.0908  & 86.0389±0.1124  & 86.0648±0.1761  & 253.0583±0.2807 \\
		&\multirow{2}{*}{ResGatedGCN}	& 4  & 152922 & 99.655±0.0777   & 86.4018±0.0764  & 86.7698±0.1167  & 256.6±1.4887    \\
		&& 16 & 350115 & 99.8654±0.0201  & 86.3926±0.1037  & 86.6744±0.1131  & 253.525±2.955   \\
		&\multirow{2}{*}{GatedGCN}	 & 4  & 152048 & 96.0368±0.0436  & 88.7754±0.0517  & \color{red}{88.9004±0.0808}  & 313.7583±3.5866 \\
		&& 16 & 352204 & 85.993±0.7327   & 83.2542±0.3806  & 83.3017±0.4354  & 800.15±15.954   \\
		&\multirow{2}{*}{GAT}   & 4  & 142659 & 99.6538±0.0963  & 86.0334±0.0754  & 86.269±0.0885   & 258.9±0.9714    \\
		&& 16 & 346947 & 99.858±0.0691   & 85.9129±0.137   & 86.3206±0.0949  & 257.55±7.1129   \\
		&\multirow{2}{*}{GIN}	& 4  & 151651 & 98.5947±0.1044  & 86.989±0.0058   & \textcolor[rgb]{ 0,  .439,  .753}{87.3379±0.0801}  & 250.5834±1.9525 \\
		&& 16 & 350427 & 95.6608±0.0557  & 87.2767±0.0569  & \textcolor[rgb]{ 0,  .439,  .753}{87.564±0.0555}   & 281.9583±2.7497 \\
		&\multirow{2}{*}{Monet}   & 4  & 150764 & 99.996±0.0058   & 87.8376±0.214   & \textcolor[rgb]{ 0,  .439,  .753}{88.0703±0.176}   & 249.875±7.1111  \\
		&& 16 & 355469 & 99.9901±0.0067  & 87.412±0.1406   & \textcolor[rgb]{ 0,  .439,  .753}{87.665±0.1235}   & 236.9083±2.5436 \\
		\bottomrule
	\end{tabular}
\end{table}

Table~\ref{tab2} shows the results of our experiments. We find that for deeper layers, the results sometimes do not improve. As the model depth increases, GNNs tend to suffer performance degeneration. The main reasons for this phenomenon are overfitting, oversmoothing~\cite{liDeeperInsightsGraph2018} and vanishing gradients. Importantly, for the Cora and CiteSeer datasets, all the GNNs clearly outperform the MLP. The results suggest that the GNNs can actually exploit the topological information of the graphs in the dataset. However, for PubMed, the power of the topological information of the graphs is inconspicuous. Due to the small datasets, the results are easily overfitted. Most GNNs reach the accuracy of the training dataset near 100\%. For all the models, GIN with 16 layers performs best in 2 small datasets and performs well in the other small dataset. For the baseline, MLP, an overly parameterized baseline, is not able to overfit the training data completely, even worsening. The baseline results have a higher standard deviation than the GNNs. This shows the training instability of the MLP.

\begin{table}[!ht]
	\caption{Performance on the SBM datasets for node classification. All the experiments were executed 4 times using the same hyperparameters with different seeds, and the mean ± standard deviation of the results was calculated. \textcolor[rgb]{ 1,  0,  0}{Red}: the best model, \textcolor[rgb]{ 0,  .439,  .753}{Blue}: good models.}
	\label{tab3}
	\footnotesize
	\centering
	\begin{tabular}{cccccccc}
		\toprule
		\textbf{dataset} & \textbf{Model}	& \textbf{L}	& \textbf{params} & \textbf{Train Acc.}	& \textbf{Val Acc.}	& \textbf{Test Acc.} & \textbf{Epoch}\\
		\midrule
		\multirow{16}[0]{*}{\begin{sideways}PATTERN\end{sideways}} & \multirow{2}[0]{*}{MLP} & 4     & 105263 & 50.2716±0.0132 & 50.501±0.0 & 50.505±0.0 & 42.0±0.0 \\
		&       & 16    & 506819 & 50.1219±0.1409 & 50.1904±0.2407 & 50.1817±0.2396 & 42.0±0.0 \\
		& \multirow{2}[0]{*}{GCN} & 4     & 100923 & 85.7572±0.038 & 85.479±0.0136 & \textcolor[rgb]{ 0,  .439,  .753}{85.6062±0.037} & 81.0±6.8313 \\
		&       & 16    & 500823 & 86.1307±0.0542 & 85.511±0.0607 & \textcolor[rgb]{ 0,  .439,  .753}{85.6337±0.0443} & 67.25±3.4034 \\
		& \multirow{2}[0]{*}{GraphSage} & 4     & 106123 & 61.5527±0.0265 & 60.7374±0.0715 & 60.8821±0.0196 & 89.5±8.8129 \\
		&       & 16    & 499887 & 63.8495±0.0858 & 62.8452±0.1194 & 62.7988±0.0528 & 92.75±11.7011 \\
		& \multirow{2}[0]{*}{ResGatedGCN} & 4     & 104003 & 84.7557±0.2293 & 84.7014±0.217 & 84.8624±0.2249 & 78.5±10.3441 \\
		&       & 16    & 502223 & 86.3522±0.0901 & 85.4158±0.0652 & \textcolor[rgb]{ 0,  .439,  .753}{85.6615±0.0625} & 63.0±10.1653 \\
		& \multirow{2}[0]{*}{GatedGCN} & 8     & 106644 & 58.8313±10.3866 & 58.8632±10.2657 & 58.8123±10.1709 & 62.25±18.3371 \\
		&       & 32    & 503339 & 75.0544±16.285 & 75.1626±16.6456 & 75.2675±16.6143 & 56.5±6.8069 \\
		& \multirow{2}[0]{*}{GAT} & 4     & 99398 & 76.9263±2.0962 & 75.1299±2.1181 & 75.1354±2.2154 & 90.5±15.9269 \\
		&       & 16    & 529806 & 93.371±0.0896 & 77.4805±0.3171 & 77.5739±0.234 & 52.5±0.5774 \\
		& \multirow{2}[0]{*}{GIN} & 4     & 100884 & 85.9095±0.0814 & 85.4469±0.0175 & \textcolor[rgb]{ 0,  .439,  .753}{85.7046±0.042} & 88.0±11.431 \\
		&       & 16    & 508574 & 85.6501±0.1001 & 85.2367±0.1151 & \textcolor[rgb]{ 0,  .439,  .753}{85.4734±0.0702} & 73.25±11.529 \\
		& \multirow{2}[0]{*}{MoNet} & 4     & 107774 & 85.4196±0.2936 & 85.303±0.2262 & \textcolor[rgb]{ 0,  .439,  .753}{85.4708±0.2121} & 94.75±23.4147 \\
		&       & 16    & 528431 & 86.2097±0.1729 & 85.6004±0.0611 & \textcolor[rgb]{ 1,  0,  0}{85.8494±0.0263} & 65.5±2.6458 \\
		\midrule
		\multirow{16}[0]{*}{\begin{sideways}CLUSTER\end{sideways}} & \multirow{2}[0]{*}{MLP} & 4     & 106015 & 20.9432±0.0016 & 20.9954±0.0045 & 20.937±0.0087 & 43.0±2.0 \\
		&       & 16    & 507691 & 20.944±0.0048 & 20.9986±0.0018 & 20.9282±0.0038 & 68.75±11.1467 \\
		& \multirow{2}[0]{*}{GCN} & 4     & 101655 & 49.743±3.5857 & 49.365±3.2667 & 48.4645±3.5681 & 63.0±4.5461 \\
		&       & 16    & 501687 & 72.5567±3.7478 & 67.4692±2.0717 & 67.6036±2.3658 & 76.0±11.7473 \\
		& \multirow{2}[0]{*}{GraphSage} & 4     & 106675 & 54.9171±0.1076 & 54.65±0.0487 & 54.4518±0.1309 & 72.75±1.7078 \\
		&       & 16    & 500503 & 63.0462±0.0499 & 62.2744±0.1038 & 62.4846±0.0369 & 80.25±11.6154 \\
		& \multirow{2}[0]{*}{ResGatedGCN} & 4     & 104355 & 62.3735±0.882 & 61.5646±0.7769 & 61.3884±0.7836 & 106.0±21.8021 \\
		&       & 16    & 502615 & 87.4099±0.9555 & 73.4934±0.1984 & \textcolor[rgb]{ 1,  0,  0}{73.56±0.292} & 60.25±3.7749 \\
		& \multirow{2}[0]{*}{GatedGCN} & 8     & 107072 & 36.3052±8.6864 & 36.4263±8.6648 & 36.1752±8.6751 & 94.25±12.5797 \\
		&       & 32    & 503911 & 20.9779±0.0587 & 20.9943±0.1439 & 20.9217±0.2132 & 45.25±2.2174 \\
		& \multirow{2}[0]{*}{GAT} & 4     & 100122 & 58.8198±0.2584 & 58.0899±0.1739 & 57.9182±0.2563 & 69.0±4.0825 \\
		&       & 16    & 530690 & 79.1966±0.4774 & 70.9993±0.2952 & \textcolor[rgb]{ 0,  .439,  .753}{71.0611±0.3584} & 70.5±9.3274 \\
		& \multirow{2}[0]{*}{GIN} & 4     & 103544 & 59.765±0.2432 & 58.397±0.1926 & 58.3287±0.1755 & 77.0±2.708 \\
		&       & 16    & 517570 & 66.4919±1.2986 & 64.736±1.1545 & 64.612±1.184 & 80.25±3.8622 \\
		& \multirow{2}[0]{*}{MoNet} & 4     & 108178 & 59.4267±0.5389 & 58.8536±0.443 & 58.6417±0.5125 & 73.25±7.719 \\
		&       & 16    & 528883 & 75.792±2.1493 & 72.7219±1.3878 & \textcolor[rgb]{ 0,  .439,  .753}{72.8281±1.3572} & 68.5±5.3229 \\
		\bottomrule
	\end{tabular}
\end{table}

Table~\ref{tab3} shows the results on the PATTERN and CLUSTERZ datasets, which are widely used in social networks. Note that they are weighted accuracies w.r.t. the class sizes, and the same training hyperparameters are used during training for a fair comparison. For PATTERN and CLUSTER, 2 classes and 6 classes exist, respectively (see Table~\ref{tab6}, Appendix A). From the baseline, MLP, in which the topological information of the graphs is not used in the models, there are only near 50\% and 21\% in the classification, just near-random guesses. Even when the layers become deeper, the results do not improve. All the GNNs except GatedGCN obtain better performance using the graphic entity. There is no overfitting in the models, and as the GNNs become deeper, the results have a large boost (except for GatedGCN, the results have a large variance when training 4 times, and when training in CLUSTER in 32 layers, the results obtain only 20.9217\%, just near the baseline). This is especially true in CLUSTER.

\begin{table}[!ht]
	\caption{Performance on the OGBN datasets including arXiv, MAG, Products and Proteins. All the results were executed 4 times using the same hyperparameters with different seeds, and the mean ± standard deviation of the results was calculated. The highest value and good values are indicated in \textcolor[rgb]{ 1,  0,  0}{red} and \textcolor[rgb]{ 0,  .439,  .753}{blue}, respectively.}
	\label{tab4}
	\scriptsize
	\centering
	\begin{tabular}{cccccccc}
		\toprule
		\textbf{dataset} & \textbf{Model}	& \textbf{L}	& \textbf{params} & \textbf{Train Acc.}	& \textbf{Val Acc.}	& \textbf{Test Acc.} & \textbf{Epoch}\\
		\midrule
		\multirow{16}[0]{*}{\begin{sideways}arxiv\end{sideways}} & \multirow{2}[0]{*}{MLP} & 4     & 87720 & 50.6525±6.5617 & 50.975±5.517 & 48.955±5.385 & 780.5±252.3496 \\
		&       & 8     & 162796 & 40.6975±15.3877 & 38.8375±20.8671 & 37.17±20.946 & 772.5±453.0 \\
		& \multirow{2}[0]{*}{GCN} & 4     & 88744 & 78.2106±0.3849 & 72.286±0.1538 & \textcolor[rgb]{ 0,  .439,  .753}{70.7924±0.0895} & 378.0±37.833 \\
		&       & 8     & 155816 & 78.7612±0.3607 & 72.5478±0.1401 & \textcolor[rgb]{ 1,  0,  0}{71.0301±0.1616} & 345.75±8.2614 \\
		& \multirow{2}[0]{*}{GraphSage} & 4     & 89435 & 72.9625±0.2958 & 70.24±0.1211 & 69.4425±0.137 & 437.75±34.2868 \\
		&       & 8     & 162775 & 72.7775±0.3179 & 70.3625±0.0685 & 69.53±0.1344 & 440.75±39.3563 \\
		& \multirow{2}[0]{*}{ResGatedGCN} & 4     & 89754 & 74.535±0.5071 & 70.3825±0.2047 & 69.23±0.1192 & 450.5±54.9272 \\
		&       & 8     & 163703 & 76.2275±0.9221 & 70.29±0.0707 & 69.23±0.1594 & 409.25±41.064 \\
		& \multirow{2}[0]{*}{GatedGCN} & 4     & 91129 & 69.135±0.8843 & 69.0125±0.5955 & 68.09±0.4032 & 780.0±285.7633 \\
		&       & 8     & 163750 & 43.985±2.7618 & 48.415±3.8031 & 49.0325±4.1766 & 251.0±77.9359 \\
		& \multirow{2}[0]{*}{GAT} & 4     & 89768 & 75.9825±0.3882 & 72.0225±0.0866 & \textcolor[rgb]{ 0,  .439,  .753}{70.6325±0.1895} & 416.0±31.0376 \\
		&       & 8     & 157864 & 77.23±0.3995 & 72.18±0.0356 & \textcolor[rgb]{ 0,  .439,  .753}{70.73±0.1512} & 411.25±33.9153 \\
		& \multirow{2}[0]{*}{GIN} & 4     & 91467 & 69.455±2.2286 & 68.3875±1.376 & 67.155±1.3146 & 851.0±166.4612 \\
		&       & 8     & 161613 & 65.0475±2.5219 & 65.0525±1.9444 & 63.93±1.9648 & 536.5±138.9208 \\
		& \multirow{2}[0]{*}{MoNet} & 4     & 91182 & 77.5175±0.4129 & 71.8275±0.2398 & \textcolor[rgb]{ 0,  .439,  .753}{70.4775±0.1473} & 397.5±24.4609 \\
		&       & 8     & 161276 & 78.3625±0.4305 & 71.9175±0.1484 & \textcolor[rgb]{ 0,  .439,  .753}{70.425±0.1586} & 336.75±18.2094 \\
		\midrule
		\multirow{16}[0]{*}{\begin{sideways}MAG\end{sideways}} & \multirow{2}[0]{*}{MLP} & 4     & 133669 & 24.31±0.3383 & 23.5375±0.2438 & 24.665±0.2271 & 261.5±46.6655 \\
		&       & 16    & 334969 & 18.3025±9.3862 & 16.8075±9.8013 & 17.4±10.7431 & 624.25±231.9258 \\
		& \multirow{2}[0]{*}{GCN} & 4     & 128605 & 31.5311±0.3482 & 29.8406±0.2038 & \textcolor[rgb]{ 0,  .439,  .753}{30.4275±0.1321} & 127.5±16.1348 \\
		&       & 16    & 329821 & 34.7873±0.5253 & 30.6906±0.0944 & \textcolor[rgb]{ 1,  0,  0}{31.1542±0.0975} & 390.5±49.4672 \\
		& \multirow{2}[0]{*}{GraphSage} & 4     & 129349 & 28.485±0.1666 & 27.31±0.2467 & 28.2925±0.2568 & 114.75±7.5443 \\
		&       & 16    & 332545 & 28.5775±0.2125 & 27.64±0.2436 & 28.62±0.2839 & 119.25±18.0069 \\
		& \multirow{2}[0]{*}{GatedGCN} & 4     & 129701 & 29.1425±0.4208 & 28.22±0.2341 & 29.3275±0.1528 & 197.75±31.3409 \\
		&       & 16    & 330749 & 25.9875±0.857 & 25.8925±0.6019 & 27.175±0.7212 & 284.5±38.3884 \\
		& \multirow{2}[0]{*}{ResGatedGCN} & 4     & 131173 & 28.7325±0.1996 & 27.4975±0.1682 & 28.3725±0.125 & 138.75±10.4363 \\
		&       & 16    & 336093 & 31.71±0.3636 & 29.715±0.3519 & \textcolor[rgb]{ 0,  .439,  .753}{30.3275±0.1674} & 479.75±80.4669 \\
		& \multirow{2}[0]{*}{GAT} & 4     & 129629 & 31.2425±0.1839 & 29.7675±0.0709 & \textcolor[rgb]{ 0,  .439,  .753}{30.4825±0.1103} & 130.25±17.0563 \\
		&       & 16    & 333917 & 31.8075±0.1438 & 30.085±0.1784 & \textcolor[rgb]{ 0,  .439,  .753}{30.7625±0.2109} & 136.75±10.2103 \\
		& \multirow{2}[0]{*}{GIN} & 4     & 130339 & 29.4775±0.2406 & 28.4675±0.1592 & 29.3925±0.2492 & 188.5±12.6886 \\
		&       & 16    & 347859 & 27.3575±0.5762 & 26.7225±0.5816 & 27.74±0.6294 & 185.75±24.0468 \\
		& \multirow{2}[0]{*}{MoNet} & 4     & 130077 & 30.715±0.5753 & 29.18±0.535 & \textcolor[rgb]{ 0,  .439,  .753}{30.195±0.4355} & 162.75±24.4046 \\
		&       & 16    & 332341 & 30.5675±0.5389 & 29.3075±0.4807 & \textcolor[rgb]{ 0,  .439,  .753}{30.1025±0.2951} & 168.25±17.595 \\
		\midrule
		\multirow{16}[0]{*}{\begin{sideways}Products\end{sideways}} & \multirow{2}[0]{*}{MLP} & 4     & 89807 & 81.015±0.6601 & 73.88±0.2464 & 59.7325±0.0378 & 77.25±5.909 \\
		&       & 16    & 300479 & 83.3975±1.3537 & 70.05±0.1252 & 56.0825±0.1723 & 78.75±6.8496 \\
		& \multirow{2}[0]{*}{GCN} & 4     & 90863 & 92.4537±0.0749 & 90.9016±0.1326 & \textcolor[rgb]{ 0,  .439,  .753}{75.2748±0.1767} & 118.5±13.3292 \\
		&       & 16    & 300299 & 92.6767±0.0047 & 91.1668±0.075 & \textcolor[rgb]{ 0,  .439,  .753}{76.018±0.1216} & 108.75±2.5 \\
		& \multirow{2}[0]{*}{GraphSage} & 4     & 92559 & 85.3525±0.0171 & 84.34±0.0408 & 66.0675±0.0842 & 93.0±6.4807 \\
		&       & 16    & 307467 & 85.3425±0.0479 & 84.4025±0.0978 & 66.5275±0.0597 & 80.5±9.5743 \\
		& \multirow{2}[0]{*}{ResGatedGCN} & 4     & 91145 & 89.225±0.1526 & 88.3125±0.1305 & 71.775±0.3205 & 154.25±17.4428 \\
		&       & 16    & 305687 & 89.05±0.4778 & 88.2325±0.4013 & 71.7875±0.7053 & 121.25±22.8674 \\
		& \multirow{2}[0]{*}{GatedGCN} & 4     & 89309 & 87.8825±0.4618 & 86.865±0.4828 & 73.355±0.4754 & 78.75±10.2429 \\
		&       & 16    & 308937 & 69.76±4.1891 & 70.64±3.6577 & 61.6875±2.0558 & 57.5±2.6458 \\
		& \multirow{2}[0]{*}{GAT} & 4     & 96879 & 91.595±0.0995 & 90.21±0.0408 & \textcolor[rgb]{ 0,  .439,  .753}{76.43±0.1485} & 148.75±12.816 \\
		&       & 16    & 291375 & 92.1325±0.096 & 90.7575±0.159 & \textcolor[rgb]{ 1,  0,  0}{77.345±0.1382} & 147.25±11.2064 \\
		& \multirow{2}[0]{*}{GIN} & 4     & 92111 & 66.2325±7.2651 & 68.9875±5.5578 & 63.2525±3.1596 & 90.0±11.7473 \\
		&       & 16    & 297655 & 16.5525±8.2426 & 22.86±8.8685 & 33.6475±5.7637 & 61.25±3.304 \\
		& \multirow{2}[0]{*}{MoNet} & 4     & 89719 & 91.93±0.2192 & 89.6875±0.1821 & 74.3725±0.0814 & 86.0±6.9762 \\
		&       & 16    & 309599 & 92.9±0.4473 & 90.085±0.3639 & \textcolor[rgb]{ 0,  .439,  .753}{75.19±0.0825} & 77.25±4.272 \\
		\midrule
		\multirow{16}[0]{*}{\begin{sideways}Proteins\end{sideways}} & \multirow{2}[0]{*}{MLP} & 4     & 89887 & 80.41±0.3601 & 75.8725±0.2874 & 70.535±0.2042 & 93.25±12.5266 \\
		&       & 16    & 301357 & 75.225±0.3854 & 71.47±0.2145 & 66.985±0.324 & 89.5±9.5743 \\
		& \multirow{2}[0]{*}{GCN} & 4     & 90967 & 85.988±0.3475 & 82.4895±0.3126 & \textcolor[rgb]{ 0,  .439,  .753}{77.9989±0.1991} & 68.25±3.2016 \\
		&       & 16    & 301204 & 86.1752±0.411 & 83.0136±0.2979 & \textcolor[rgb]{ 0,  .439,  .753}{76.4456±0.7319} & 71.25±12.842 \\
		& \multirow{2}[0]{*}{GraphSage} & 4     & 89978 & 76.8275±0.0457 & 69.5225±0.1848 & 68.71±0.0648 & 52.5±3.0 \\
		&       & 16    & 304967 & 77.375±0.09 & 69.7525±0.1367 & 68.95±0.1465 & 57.0±3.9158 \\
		& \multirow{2}[0]{*}{ResGatedGCN} & 4     & 89509 & 84.87±0.5857 & 81.4275±0.6523 & \textcolor[rgb]{ 0,  .439,  .753}{78.18±0.8586} & 55.25±7.932 \\
		&       & 16    & 304132 & 87.635±0.2428 & 83.765±0.1287 & \textcolor[rgb]{ 1,  0,  0}{79.7175±0.2037} & 229.75±17.1731 \\
		& \multirow{2}[0]{*}{GatedGCN} & 4     & 90803 & 72.0325±3.3052 & 65.6975±7.5995 & 62.4425±8.421 & 45.5±1.7321 \\
		&       & 16    & 305897 & 58.11±4.0229 & 53.4275±4.686 & 53.7925±1.1951 & 54.5±3.3166 \\
		& \multirow{2}[0]{*}{GAT} & 4     & 93272 & 83.4925±0.6073 & 79.8625±0.5536 & \textcolor[rgb]{ 0,  .439,  .753}{75.1025±1.388} & 46.25±3.2016 \\
		&       & 16    & 323384 & 85.4675±0.4718 & 81.5425±0.4513 & \textcolor[rgb]{ 0,  .439,  .753}{77.03±0.6224} & 45.5±2.3805 \\
		& \multirow{2}[0]{*}{GIN} & 4     & 90039 & 47.395±5.8131 & 46.1875±6.8098 & 47.26±5.1833 & 44.25±1.893 \\
		&       & 16    & 299350 & 49.04±2.5863 & 49.12±2.7496 & 49.36±2.0184 & 46.5±1.7321 \\
		& \multirow{2}[0]{*}{MoNet} & 4     & 92704 & 82.8725±1.2751 & 79.7775±1.0561 & 74.8025±1.0658 & 45.75±2.5 \\
		&       & 16    & 299019 & 83.3425±1.303 & 79.5875±1.3526 & \textcolor[rgb]{ 0,  .439,  .753}{75.06±0.8849} & 49.25±9.9121 \\

		\bottomrule
	\end{tabular}
\end{table}

\begin{table}[!ht]
	\caption{Performance on the OGBN datasets including arXiv and Products. All the experiments were executed 4 times using the same hyperparameters with different seeds, and the mean ± standard deviation of the results was calculated. The indicators that improved significantly are highlighted in \textcolor[rgb]{ 1,  0,  0}{red}, while the merely improved indicators are highlighted in \textcolor[rgb]{ 0,  .439,  .753}{blue}. Models with the suffix -pe used Laplacian eigenvectors as node positional encodings with a dimension of 64. Models with the suffix -ne used node2vec embeddings in the input features.}
	\label{tab5}
	\centering
	\footnotesize
	\begin{tabular}{cccccccc}
		\toprule
		\textbf{dataset} & \textbf{Model}	& \textbf{L}	& \textbf{params} & \textbf{Train Acc.}	& \textbf{Val Acc.}	& \textbf{Test Acc.} & \textbf{Epoch}\\
		\midrule
		\multirow{18}[0]{*}{\begin{sideways}arxiv\end{sideways}} & \multirow{2}[0]{*}{MLP} & 4     & 87720 & 50.6525±6.5617 & 50.975±5.517 & 48.955±5.385 & 780.5±252.3496 \\
		&       & 8     & 162796 & 40.6975±15.3877 & 38.8375±20.8671 & 37.17±20.946 & 772.5±453.0 \\
		& \multirow{2}[0]{*}{MLP-pe} & 4     & 96040 & 36.5803±15.0763 & 34.8024±19.657 & 32.6374±19.626 & 776.25±445.5 \\
		&       & 8     & 171376 & 25.7848±15.7575 & 18.1466±21.0376 & 16.5201±21.3166 & 253.25±289.8521 \\
		& \multirow{2}[0]{*}{MLP-ne} & 4     & 104104 & 72.4907±0.4172 & 69.8774±0.1538 & \textcolor[rgb]{ 1,  0,  0}{68.879±0.1818} & 667.5±71.9838 \\
		&       & 8     & 179692 & 70.3805±2.2619 & 65.8±1.2338 & \textcolor[rgb]{ 1,  0,  0}{64.5999±0.9095} & 757.5±185.5344 \\
		& \multirow{2}[0]{*}{GCN} & 4     & 88744 & 78.2106±0.3849 & 72.286±0.1538 & 70.7924±0.0895 & 378.0±37.833 \\
		&       & 8     & 155816 & 78.7612±0.3607 & 72.5478±0.1401 & 71.0301±0.1616 & 345.75±8.2614 \\
		& \multirow{2}[0]{*}{GCN-pe} & 4     & 97064 & 77.8769±0.3842 & 72.1467±0.1701 & \textcolor[rgb]{ 0,  .439,  .753}{70.8665±0.1229} & 358.5±11.7898 \\
		&       & 8     & 164136 & 78.9484±0.1457 & 72.4572±0.1321 & 70.9961±0.0608 & 336.25±14.3614 \\
		& \multirow{2}[0]{*}{GCN-ne} & 4     & 105128 & 78.8789±0.1964 & 73.1568±0.094 & \textcolor[rgb]{ 0,  .439,  .753}{71.814±0.1} & 310.75±14.2916 \\
		&       & 8     & 172200 & 79.5466±0.2921 & 73.0906±0.1756 & \textcolor[rgb]{ 0,  .439,  .753}{71.7137±0.0579} & 306.25±16.2763 \\
		& \multirow{2}[0]{*}{GAT} & 4     & 89768 & 75.9825±0.3882 & 72.0225±0.0866 & 70.6325±0.1895 & 416.0±31.0376 \\
		&       & 8     & 157864 & 77.23±0.3995 & 72.18±0.0356 & 70.73±0.1512 & 411.25±33.9153 \\
		& \multirow{2}[0]{*}{GAT-pe} & 4     & 98088 & 75.5853±0.5619 & 71.8078±0.1395 & \textcolor[rgb]{ 0,  .439,  .753}{70.651±0.2231} & 387.0±24.3721 \\
		&       & 8     & 166184 & 77.2756±0.7433 & 72.0561±0.0529 & 70.6772±0.0362 & 353.0±21.9089 \\
		& \multirow{2}[0]{*}{GAT-ne} & 4     & 106152 & 77.5594±0.4013 & 72.9874±0.1534 & \textcolor[rgb]{ 0,  .439,  .753}{71.5666±0.0926} & 361.0±11.6333 \\
		&       & 8     & 174248 & 78.0253±0.2107 & 72.7978±0.1711 & \textcolor[rgb]{ 0,  .439,  .753}{71.2682±0.1982} & 322.25±3.304 \\
		\midrule
		\multirow{18}[0]{*}{\begin{sideways}Products\end{sideways}} & \multirow{2}[0]{*}{MLP} & 4     & 89807 & 81.015±0.6601 & 73.88±0.2464 & 59.7325±0.0378 & 77.25±5.909 \\
		&       & 16    & 300479 & 83.3975±1.3537 & 70.05±0.1252 & 56.0825±0.1723 & 78.75±6.8496 \\
		& \multirow{2}[0]{*}{MLP-pe} & 4     & 98387 & 81.2832±0.3376 & 73.9917±0.262 & \textcolor[rgb]{ 0,  .439,  .753}{59.7828±0.0457} & 80.25±4.272 \\
		&       & 16    & 309059 & 83.1328±0.3495 & 69.9654±0.5337 & 55.8319±0.4602 & 78.5±5.1962 \\
		& \multirow{2}[0]{*}{MLP-ne} & 4     & 106703 & 93.7025±0.0996 & 89.2162±0.086 & \textcolor[rgb]{ 1,  0,  0}{70.9476±0.3534} & 68.5±2.3805 \\
		&       & 16    & 317375 & 92.9284±1.5731 & 87.5626±0.2582 & \textcolor[rgb]{ 1,  0,  0}{66.9452±0.446} & 68.0±5.7155 \\
		& \multirow{2}[0]{*}{GCN} & 4     & 90863 & 92.4537±0.0749 & 90.9016±0.1326 & 75.2748±0.1767 & 118.5±13.3292 \\
		&       & 16    & 300299 & 92.6767±0.0047 & 91.1668±0.075 & 76.018±0.1216 & 108.75±2.5 \\
		& \multirow{2}[0]{*}{GCN-pe} & 4     & 99443 & 92.4556±0.1256 & 90.8343±0.1218 & \textcolor[rgb]{ 0,  .439,  .753}{75.3129±0.1286} & 127.5±13.0767 \\
		&       & 16    & 308814 & 92.7685±0.1377 & 91.2646±0.1161 & \textcolor[rgb]{ 0,  .439,  .753}{76.1586±0.1262} & 121.75±9.7767 \\
		& \multirow{2}[0]{*}{GCN-ne} & 4     & 107759 & 93.5741±0.1473 & 91.7135±0.0884 & 75.0002±0.3371 & 85.0±10.4243 \\
		&       & 16    & 317067 & 93.6397±0.1337 & 91.772±0.0534 & 74.6966±0.2146 & 94.75±9.4296 \\
		& \multirow{2}[0]{*}{GAT} & 4     & 96879 & 91.595±0.0995 & 90.21±0.0408 & 76.43±0.1485 & 148.75±12.816 \\
		&       & 16    & 291375 & 92.1325±0.096 & 90.7575±0.159 & 77.345±0.1382 & 147.25±11.2064 \\
		& \multicolumn{1}{c}{\multirow{2}[0]{*}{GAT-pe}} & 4     & 105719 & 91.5923±0.0928 & 90.3072±0.0941 & \textcolor[rgb]{ 0,  .439,  .753}{76.4812±0.1802} & 156.75±16.5202 \\
		&       & 16    & 299695 & 92.3608±0.0987 & 90.8387±0.1179 & \textcolor[rgb]{ 0,  .439,  .753}{77.3693±0.3204} & 144.0±4.5461 \\
		& \multirow{2}[0]{*}{GAT-ne} & 4     & 114287 & 93.2509±0.1768 & 91.4738±0.0939 & 76.2792±0.2283 & 104.25±11.8708 \\
		&       & 16    & 307759 & 93.4481±0.2017 & 91.6169±0.1513 & 76.259±0.3628 & 116.0±10.2307 \\
		
		\bottomrule
	\end{tabular}
\end{table}

Unlike the datasets for node classification, which have only 1 graph, PATTERN and CLUSTER have 14k and 12k graphs, respectively. The task is to classify the nodes in each graph. It seems that for these tasks, oversmoothing is not a serious problem because not all the nodes are connected together. It can also be found in the results that the accuracy of the classification improves when the layers become deeper. MoNet with 16 layers achieves the best performance.

The OGBN dataset in Table~\ref{tab4} was collected entirely from the Open Graph Benchmark~\cite{huOpenGraphBenchmark2020a}. The recent minibatch-based GNNs are always used in OGBN because there is only 1 graph, and these graphs can easily exceed the available CUDA memory during training. Before training using the different model architectures, we first randomly sample the nodes within a graph and return their induced subgraph to separate the dataset for all the datasets except arXiv (for arXiv, when choosing the appropriate parameters, the CUDA memory is sufficient to address the full-batch version of GNNs).

We can find that both GNNs perform well on arXiv and MAG, which both come from academic graphs; however, because MAG includes more classes than arXiv, performing classification in MAG is more difficult.
Among these 4 datasets, the models GAT and GCN perform well compared to the others, while GIN, which uses feature outputs from each layer of the network, performs the worst---even worse than the baseline in Products when using 16 layers. As the layers become deeper, the MLP degrades on all 4 datasets. It seems that for the baseline, a large parameter budget does not achieve better performance.

Numerous methods for data augmentation have been proposed in the image classification field that can increase the classification accuracy. Here, we adopt the Node2vec embeddings and Laplacian eigenvectors to investigate how the input features affect the accuracy (see Table~\ref{tab5}). The object datasets, including arXiv and Products, are used to test how the input features affect accuracy with the MLP, GCN and GAT models. We use MLP as the baseline and include GCN and GAT because all these models perform well on these 2 tasks. MLP in particular achieves a substantial improvement on 2 of the datasets with the node embedding features. It appears that the node embedding well represents the topological information in the input features, causing a large effect. When using the node embedding on arXiv, the GNNs still achieve improvements; it seems that input features containing the topological information still have effects on arXiv. However, when the GNNs are applied to the Products dataset, no improvements are found. All the models with Laplacian eigenvectors (except MLP with 16 layers) achieve improvements on Products, but only GCN and GAT with 4 layers improve on arXiv. This phenomenon shows that the impact of the location information described by the Laplacian eigenvectors plays a larger role in the Products dataset than in the arXiv dataset.


%
%
%

\section{Conclusions}

This study created a benchmark framework for node classification with GNNs, allowing researchers to test how the architectures and input features of various models affect the results. A k-fold model assessment was defined and applied to the small datasets. We also defined a set of model training procedures, including how to decrease the learning rate and when to terminate training. The result is a standard experimental pipeline for GNNs that helps ensure fair model comparisons.
For all the models, 2 parameter budgets with different numbers of layers are used to test whether the number of parameters affects the results. Overall, our goal in this study was to construct a benchmark to assist in testing model performance on different datasets instead of obtaining the best classification accuracy.
\begin{itemize}
	\item	We found that for graphs containing topological information, all the GNNs performed well compared with the baseline on the small datasets except for PubMed. It appears that topological information does not play an important role in improving GNN classification accuracy on PubMed.
	\item	For the artificially generated SBMs used in social networks, we found that increasing the number of layers elevates the results, especially on CLUSTER for most models. It appears that oversmoothing is not a serious problem in node classification tasks on datasets that contain many unconnected graphs.
	\item	To accommodate CUDA memory limitations, we first randomly sample nodes within a graph and return their induced subgraph to separate the dataset into OGBN datasets (e.g., MAG, Products and Proteins). The results show that GCN and GAT perform best on the OGBN datasets and indicate that the traditional classical models are still highly useful.
	\item	We applied some data augmentation methods in our experiments, including node2vec embeddings and Laplacian eigenvectors. We found that the use of node2vec embeddings in MLP provides a substantial improvement on the arXiv and Products datasets and improved the GNN performance on arXiv but did not work with GNNs on the Products dataset. In contrast, the Laplacian eigenvectors did help on Products, but the improvement was limited.
\end{itemize}


\subsection{Future Work}
Methods such as dropedge~\cite{rongDropedgeDeepGraph2019a}, pairnorm~\cite{zhaoPairnormTacklingOversmoothing2019a} and nodenorm~\cite{zhouEffectiveTrainingStrategies2020a} have been proposed to solve the problems of oversmoothing and overfitting. However, these methods were tested only on small datasets in the original studies. Thus, their applicability to large datasets has not been fully verified. The experimental conditions with respect to the split among training/validation/testing and model assessment do not seem to ensure a fair comparison. Therefore, in future work, we will try to use our benchmark to test the role of the above methods in reducing the convergence speed, oversmoothing, and overfitting.
When working with directed graphs, we first add edges to make them undirected, and it will be fruitful to explore how edge direction information can be considered to improve prediction performance. In addition, some datasets include some node temporal information (e.g., the year in which papers are published in arXiv and MAG); this temporal information could be included in the input features to improve the classification accuracy.
To accommodate CUDA memory limitations, for some datasets, we only randomly sampled the graph nodes and returned an induced subgraph to partition the dataset. Some minibatch GNNs, especially NeighborSampling~\cite{hamiltonInductiveRepresentationLearning2017}, GraphSAINT~\cite{zengGraphsaintGraphSampling2019} and ClusterGCN~\cite{chiangClusterGCNEfficientAlgorithm2019}, are used to perform node classification. Sometimes, these even slightly outperform the full-batch version of GNNs, which does not fit into ordinary GPU memory. Our benchmark framework is also suitable for testing these models.

\appendix
\unskip
\section{}\label{appA}
\unskip
\begin{figure}[!ht]
	\centering
	\includegraphics[width=15.6 cm]{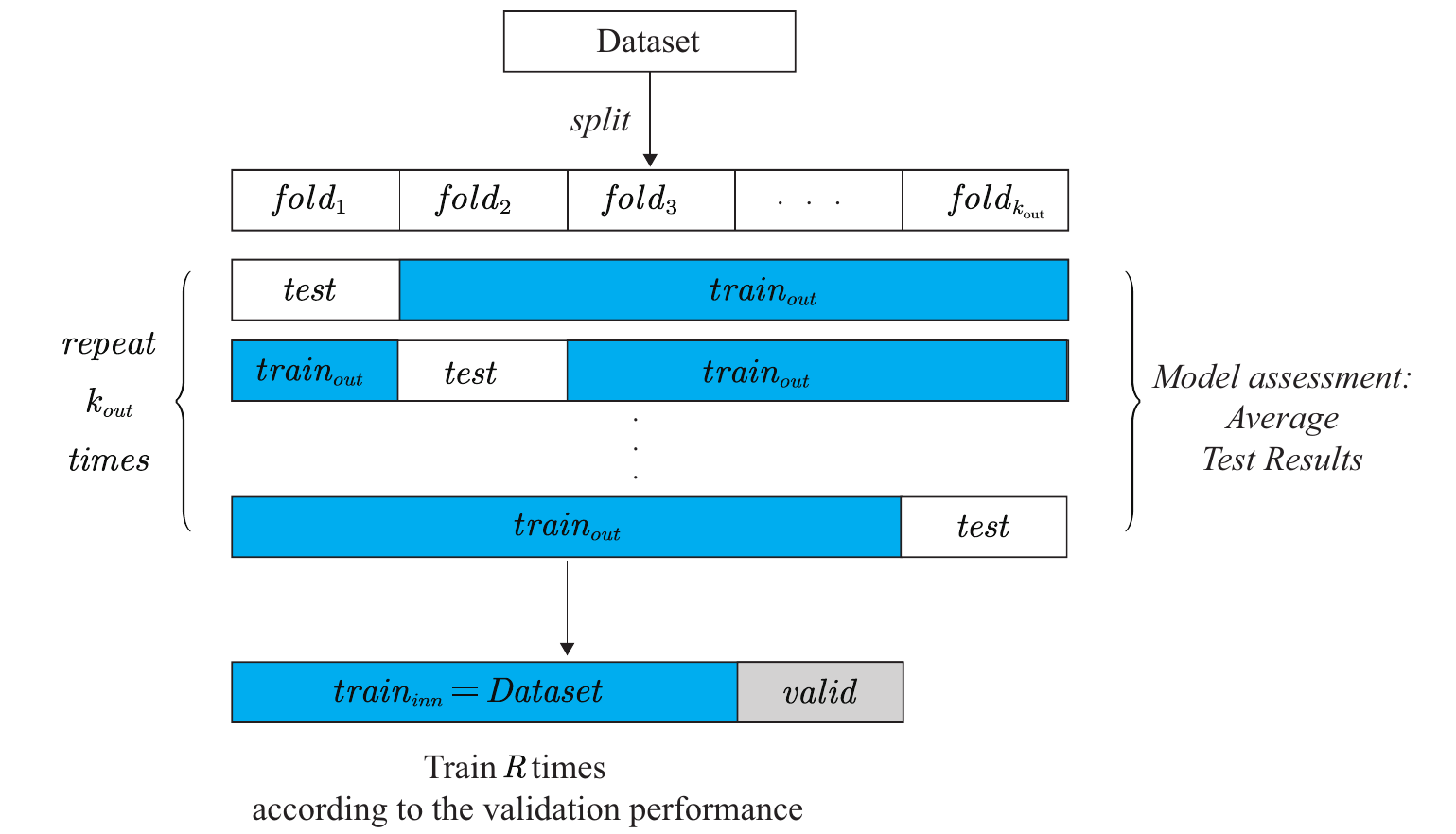}
	\caption{The evaluation framework for small datasets}
	\label{fig4}
\end{figure}
\unskip
\begin{figure}[!ht]
	\centering
	\includegraphics[width=15.6 cm]{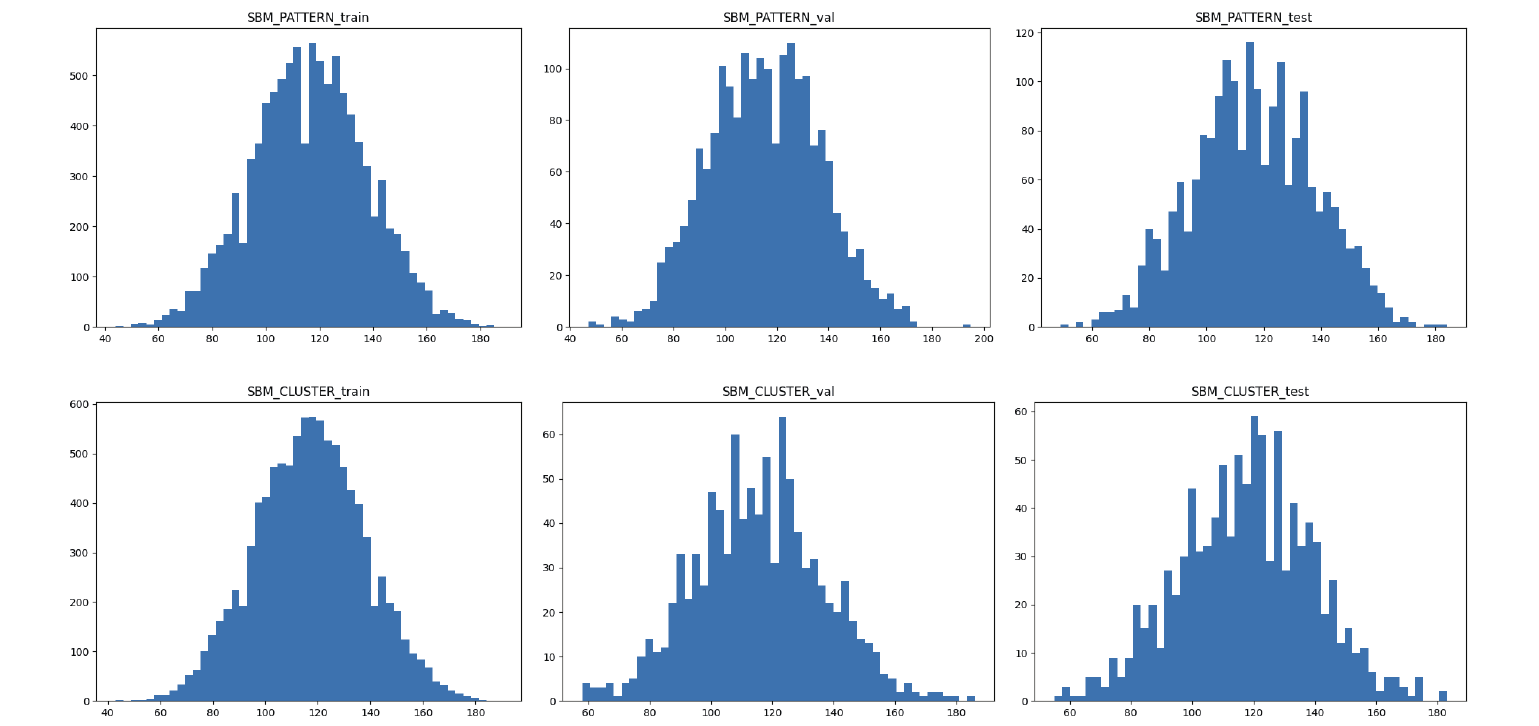}
	\caption{A histogram of the number of nodes in each graph in the PATTERN and CLUSTER datasets}
	\label{fig5}
\end{figure}
\unskip
\begin{table}[!ht]
	\caption{Summary of the datasets used in experiments. A split strategy, prediction task and input features are described for each dataset; more details can be found in Section 2.1.}
	\label{tab6}
	\centering
	\footnotesize
	\begin{tabular}{cccccc}
		\toprule
		\textbf{Domain}	& \textbf{Name }	& \textbf{\#graphs} & \textbf{\#nodes*dim}	& \textbf{\#edges*dim}	& \textbf{Classes}\\
		\midrule
		\multirow{3}{*}{citation network datasets}                 & Cora     & 1        & 2708*1433   & 5278              & 7       \\
		& CiteSeer & 1        & 3327*3703   & 4552              & 6       \\
		& PubMed   & 1        & 19,717*500  & 44,324            & 3       \\
		\multirow{2}{*}{Mathematical Modelling in social networks} & PATTERN  & 14k      & 44-195*1    & 752$\sim$15900    & 2       \\
		& CLUSTER  & 12k      & 43-190*1    & 524$\sim$10752    & 6       \\
		\multirow{2}{*}{Academic graphs}                           & arXiv    & 1        & 169343*128  & 1166243(directed) & 40      \\
		& MAG      & 1        & 736389*128  & 5416271(directed) & 349     \\
		Commercial networks                                     & Products & 1        & 2449029*100 & 123718280         & 47      \\
		Biological networks                                    & Proteins & 1        & 132,534     & 39561252*8        & 2  \\
		\bottomrule
	\end{tabular}
\end{table}


\section{}\label{appB}
\begin{figure}[!ht]
	\centering
	\includegraphics[width=15.6 cm]{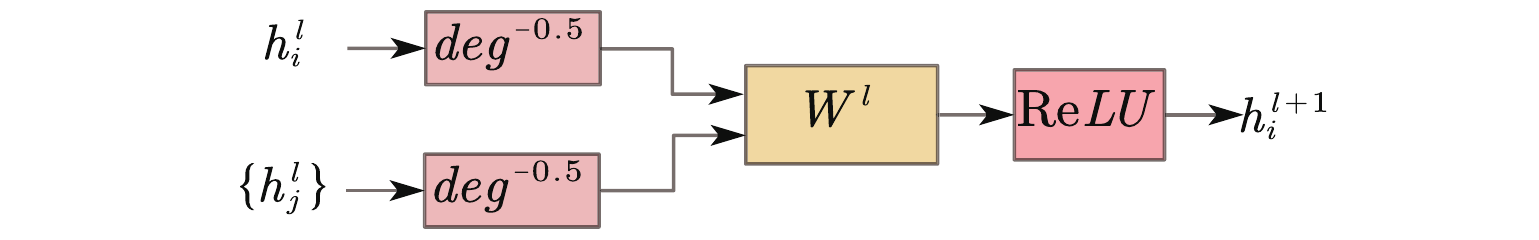}
	\caption{GCN Layer}
	\label{fig6}
\end{figure}

\unskip
GCN:
Mathematically, the GCN model follows this formula:
\begin{equation}
h^{(l+1)}=\sigma\left(\tilde{D}^{-\frac{1}{2}} \tilde{A} \tilde{D}^{-\frac{1}{2}} h^{(l)} W^{(l)}\right) 
\end{equation}

where $H^{(l)}$ denotes the $l^{(th)}$ layer in the network, $\sigma$ is the nonlinearity, and $W$ is the weight matrix for this layer. $\tilde{D}^{-\frac{1}{2}} \tilde{A} \tilde{D}^{-\frac{1}{2}}$ indicates a renormalization trick in which there is a self-connection to each node of the graph. Therefore, $\tilde{D}$ is the corresponding degree matrix of $A+I$, and $\tilde{A}= A+I$. The shape of $H^{(0)}$ is $N \times D$, where $N$ is the number of nodes, and $D$ is the number of input features.
For better understanding, the following similar formula can also be used to describe the models.
\begin{equation}
h_i^{(l+1)} = \sigma \left(\sum_{j \in \mathcal{N}(i) \cup \{ i \}} \frac{1}{\sqrt{\deg(i)} \cdot \sqrt{\deg(j)}} \cdot (h_j^{(l)} W^{(l)})\right)  
\end{equation}

where $\sqrt{\deg(i)}$ and $\sqrt{\deg(j)}$ are the degrees of nodes $i$ and $j$, respectively; $\mathcal{N}(i)$ is a neighbor of node $i$; and $\sigma$ is the activation function; here, we used ReLU as the activation function (see Figure~\ref{fig6}).

\begin{figure}[!ht]
	\centering
	\includegraphics[width=15.6 cm]{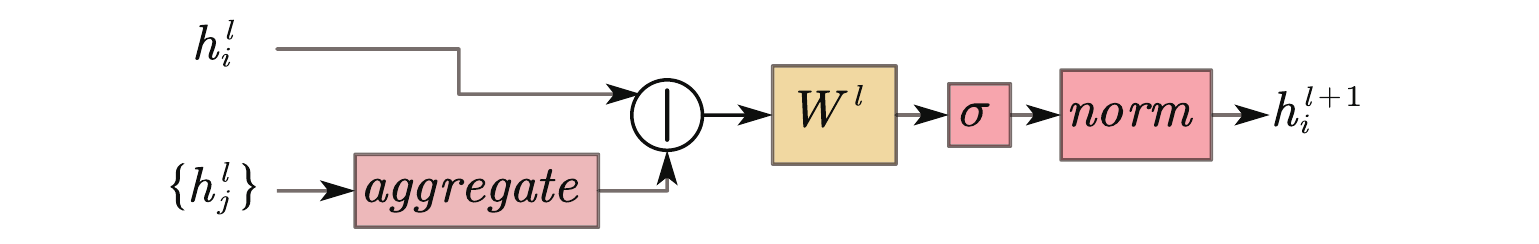}
	\caption{GraphSage Layer}
	\label{fig7}
\end{figure}

GraphSage:
GraphSage acts as a framework for aggregating information about adjacent nodes. Under this framework, aggregate functions can be used to combine information from adjacent nodes. We then can access information from adjacent nodes when processing the current node. Finally, a norm is used for the vector that combines the information of the current node and adjacent nodes.
\begin{equation}
h_{\mathcal{N}(i)}^{(l+1)} =\operatorname{aggregate}\left(\left\{h_{j}^{l}, \forall j \in \mathcal{N}(i)\right\}\right)
\end{equation}
\begin{equation}
h_{i}^{(l+1)} =\sigma\left(\operatorname{concat}\left(h_{i}^{l}, h_{\mathcal{N}(i)}^{l+1}\right) \cdot W^{l}\right)
\end{equation}
\begin{equation}
h_{i}^{(l+1)} =\operatorname{norm}\left(h_{i}^{l+1}\right)
\end{equation}

where node $j$ is a neighbor of $i$, and $\sigma$ is the activation function (see Figure~\ref{fig7}).

\begin{figure}[!ht]
	\centering
	\includegraphics[width=15.6 cm]{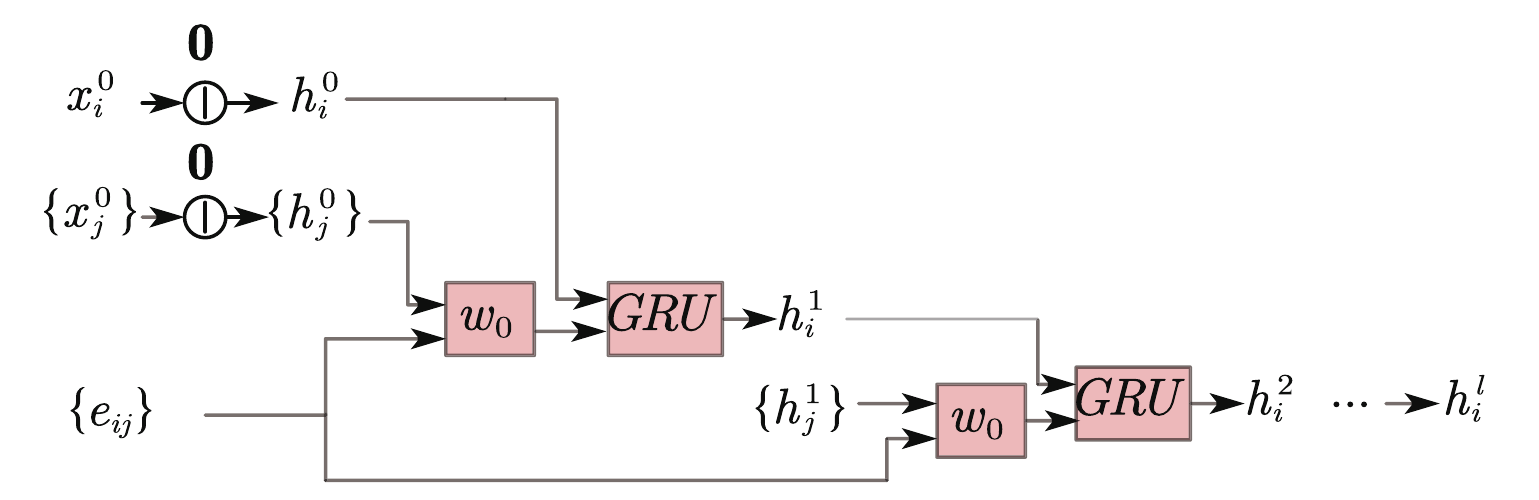}
	\caption{GatedGCN layer}
	\label{fig8}
\end{figure}

GatedGCN:
In the formula, the number of input channels of $\mathbf{x}_{i}$ must be less than or equal to the number of output channels. If there are fewer input channels than output channels, we first use zero vectors to complete the input channels to obtain $\mathbf{h}_{i}^{(0)}$ to make the input channels equal to the output channels. $\Theta$ is the learnable parameter, and $e_{ij}$ is the edge weight. Finally, a gated recurrent unit (GRU) is used in the algorithm to implement long-term memory (see Figure~\ref{fig8}).
\begin{equation}
\mathbf{h}_{i}^{(0)} =\mathbf{x}_{i} \| \mathbf{0}
\end{equation}
\begin{equation}
\mathbf{m}_{i}^{(l+1)} =\sum_{j \in \mathcal{N}(i)} \Theta \cdot e_{ij} \cdot \mathbf{h}_{j}^{(l)}
\end{equation}
\begin{equation}
\mathbf{h}_{i}^{(l+1)} =\operatorname{GRU}\left(\mathbf{m}_{i}^{(l+1)}, \mathbf{h}_{i}^{(l)}\right)
\end{equation}

Unlike the GCN, which adds the surrounding information, and GraphSAGE, which concatenates the surrounding information into its own embedding, GatedGCN uses the GRU method to collect the surrounding information and support modeling of long-term dependencies.

\begin{figure}[!ht]
	\centering
	\includegraphics[width=0.8\textwidth]{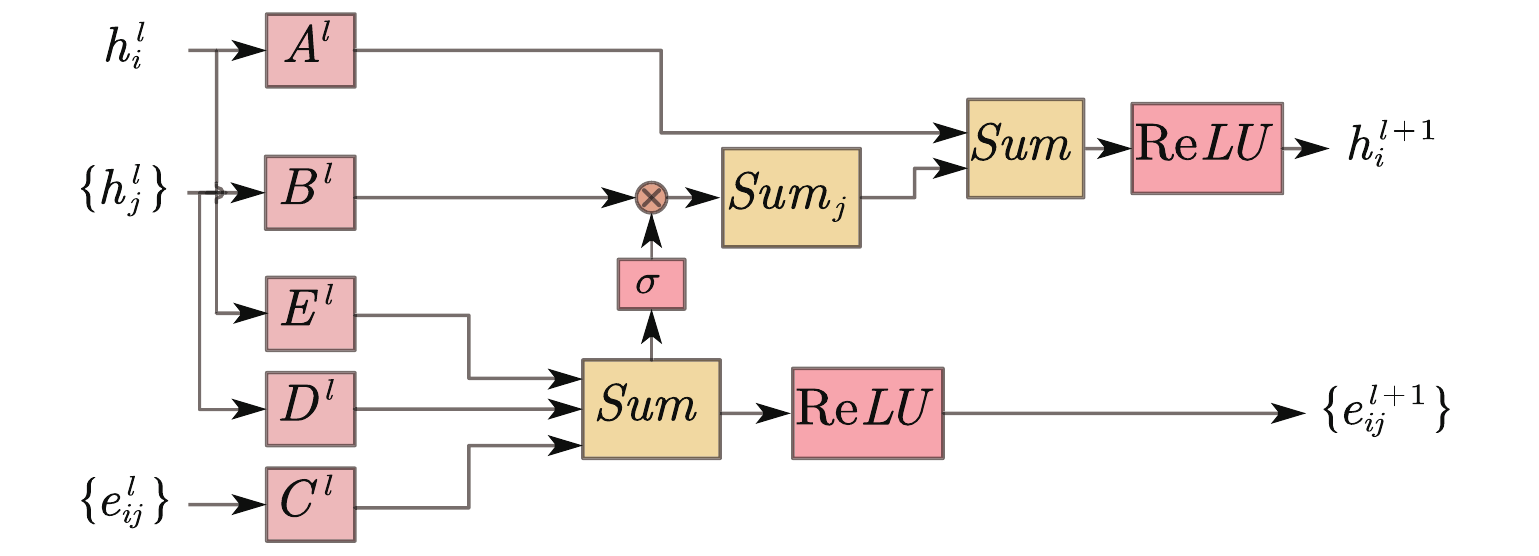}
	\caption{ResGatedGCN layer}
	\label{fig9}
\end{figure}

ResGatedGCN:
In the ResGatedGCN formula, $A^l,B^l,C^l,D^l,E^l$ are all learnable parameters, $\varepsilon$ is a small fixed constant for numerical stability, and$\sigma$ is the sigmoid function.
In detail, the edge gates~\cite{duvenaudConvolutionalNetworksGraphs2015} can be regarded as a soft attention process, which is related to the standard sparse attention mechanism~\cite{bahdanauNeuralMachineTranslation2014}. In contrast to other GNNs, the GatedGCN architecture explicitly maintains the edge features $e_{ij}$ in each layer~\cite{bressonTwostepGraphConvolutional2019}~\cite{joshiEfficientGraphConvolutional2019} (see Figure~\ref{fig9}).
\begin{equation}
\hat{e}_{i j}^l = E^{l} h_{i}^{l}+B^{l} h_{j}^{l}+C^{l} {e}_{i j}^{l}
\end{equation}
\begin{equation}
{e}_{i j}^{l+1}= ReLU(\hat{e}_{i j}^l)
\end{equation}
\begin{equation}
h_i^{l+1}=ReLU\left(A^{l}h_{i}^{l}+\frac{\sum_{j \in \mathcal{N}}\sigma\left(\hat{e}_{i j}^{l}\right)B^{l}h_j^{l}}{\sum_{j \in \mathcal{N}} \sigma\left(\hat{e}_{i j}^{l}\right)+\varepsilon} \right)
\end{equation}

\begin{figure}[!ht]
	\centering
	\includegraphics[width=0.8\textwidth]{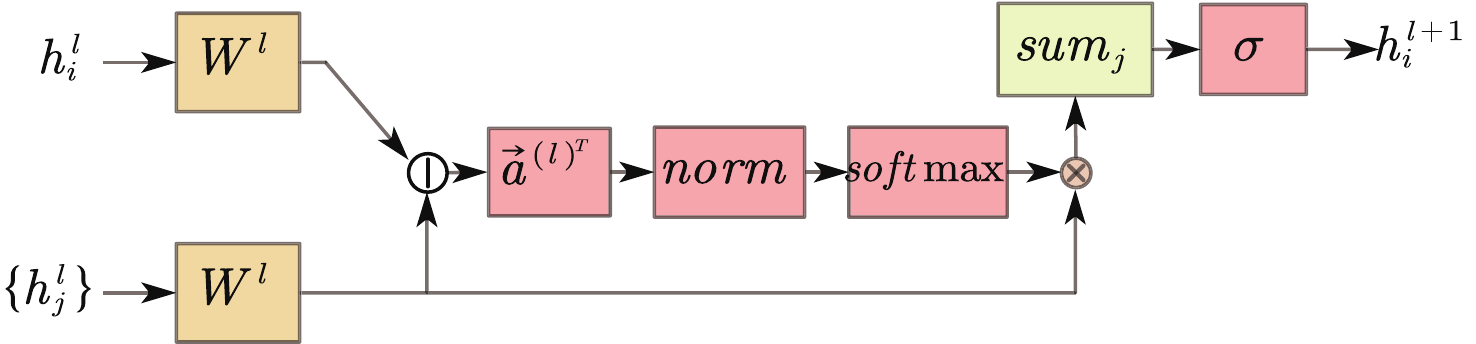}
	\caption{GAT layer}
	\label{fig10}
\end{figure}

GAT:
The core idea of the GAT is to learn the importance of neighboring nodes and then use the learned importance weight to carry out weighted summation to update a node's own embedding.
\begin{equation}
z_{i}^{(l)} =W^{(l)} h_{i}^{(l)}
\end{equation}
\begin{equation}
e_{i j}^{(l)} =\operatorname{LeakyReLU}\left(\vec{a}^{(l)^{T}}\left(z_{i}^{(l)} \| z_{j}^{(l)}\right)\right) 
\end{equation}
\begin{equation}
\alpha_{i j}^{(l)} =\frac{\exp \left(e_{i j}^{(l)}\right)}{\sum_{k \in \mathcal{N}(i)} \exp \left(e_{i k}^{(l)}\right)}
\end{equation}
\begin{equation}
h_{i}^{(l+1)} =\sigma\left(\sum_{j \in \mathcal{N}(i)\cup \{ i \}} \alpha_{i j}^{(l)} z_{j}^{(l)}\right)
\end{equation}
where $[\|]$ indicates concatenating the features of node $i$ and node $j$, which have been added by the linear map of the shared parameter $W$. $\vec{a}^{T}(.)$ is a learnable weight vector. A single-layer feedforward neural network is applied; then, $\text { LeakyReLU }(.)$ and $\operatorname{softmax}_{i}(.)$ are used to calculate the attention coefficient $\alpha_{i j}^{l}$. Finally, the node embedding is aggregated using the learned coefficient $\alpha_{i j}^{l}$ to obtain the node embedding of the $l+1$ layer (see Figure~\ref{fig10}).

\begin{figure}[!ht]
	\centering
	\includegraphics[width=0.8\textwidth]{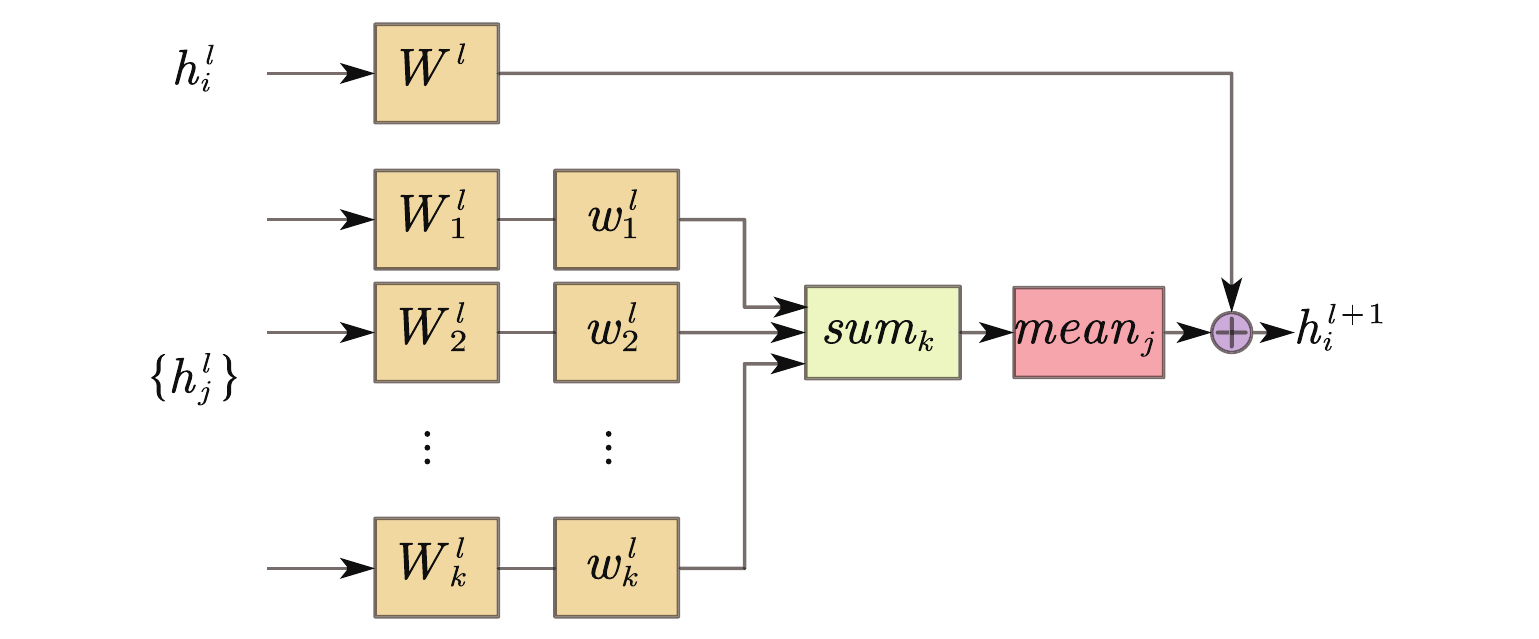}
	\caption{Monet layer}
	\label{fig11}
\end{figure}

MONET:
Mixture model networks are a deep network framework run in non-Euclidean space. In the case of graphs, the node update equation is defined as follows:

\begin{equation}
z_{i}^{(l)} =W^{(l)} h_{i}^{(l)}
\end{equation}
\begin{equation}
h_{i}^{(l+1)}  = \frac{1}{|\mathcal{N}(i)|}
\sum_{j \in \mathcal{N}(i)} \sum_{k=1}^K
\mathbf{w}_k(\mathbf{e}_{i,j}) \odot \mathbf{\Theta}_k h_j^l + h_{i}^{(l)}\mathbf{\Theta} 
\end{equation}
\begin{equation}
\mathbf{w}_k(\mathbf{e}) = \exp \left( -\frac{1}{2} {\left(
	\mathbf{e} - \mathbf{\mu}_k \right)}^{\top} \Sigma_k^{-1}
\left( \mathbf{e} - \mathbf{\mu}_k \right) \right) 
\end{equation}
\begin{equation}
\mathbf{e}=\tanh \left(\mathbf{\Theta}\left(\operatorname{deg}_{i}^{-1 / 2}, \operatorname{deg}_{j}^{-1 / 2}\right)^{T}\right)
\end{equation}

where $\mu_{k}$ and $\Sigma_{k}^{-1}$ denote the (learnable) parameters of the mean vector and diagonal covariance matrix, respectively; $\mathbf{\Theta}_k$ and $\mathbf{\Theta}$ are the learnable parameters; and $\mathbf{e}$ and $\mathbf{e}_{ij}$ are the same. The edge attributes of nodes $i$ and $j$ are shown in Figure~\ref{fig11}.

\begin{figure}[!ht]
	\centering
	\includegraphics[width=15.6 cm]{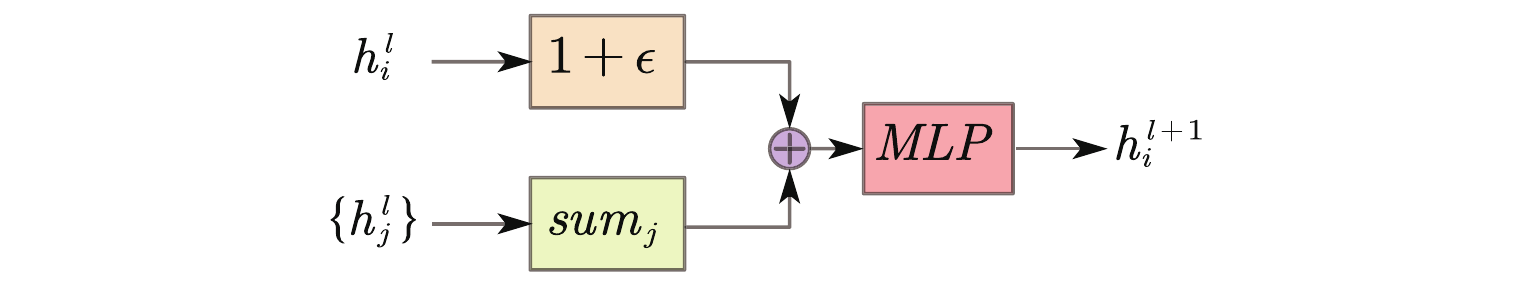}
	\caption{GIN layer}
	\label{fig12}
\end{figure}

GIN:
The equations below compute the node embedding $h_{i}^{(l+1)}$ of layer $l+1$ from the embeddings of layer $l$:
\begin{equation}
h_{i}^{(l+1)}=\operatorname{MLP}^{(l)}\left(\left(1+\epsilon^{(l)}\right) \cdot h_{i}^{(l)}+\sum_{j \in N(i)} h_{i}^{(l)}\right) 
\end{equation}

where $l$ denotes the $l$-th layer in the network, $\epsilon$ is a learnable parameter or a fixed scalar, $N(i)$ denotes the set of neighbor indices of node $i$, and $MLP$ represents the multilayer perceptron(see Figure~\ref{fig12}). In the first iteration, if the input features are one-hot encodings, $MLP$ is not used before summation. For GIN, we consider the feature outputs from each layer of the network.


\bibliographystyle{elsarticle-num} 
\bibliography{references.bib}


%
%
%
\end{document}